%% file: iclr2027_conference.tex
\documentclass{article} % For LaTeX2e
\usepackage{iclr2027_conference,times}

\input{math_commands.tex}

\usepackage[hypertexnames=false]{hyperref}
\usepackage{url}
\usepackage{graphicx}
\usepackage{placeins}
\usepackage{booktabs}
\usepackage{multirow}
\usepackage{colortbl}
\usepackage{listings}
\usepackage{float}
\usepackage{fontawesome5}
\usepackage{enumitem}

\usepackage{array,tabularx}
\usepackage[table]{xcolor}
\usepackage[most]{tcolorbox}

\definecolor{promptframe}{RGB}{122,142,172}
\definecolor{prompttitle}{RGB}{214,226,240}
\definecolor{promptbody}{RGB}{247,249,252}

\newtcolorbox{promptbox}[1]{%
  enhanced, breakable,
  colback=promptbody, colframe=promptframe,
  colbacktitle=prompttitle, coltitle=black,
  title={\sffamily\bfseries #1},
  boxrule=0.7pt, arc=1.5mm,
  left=8pt, right=8pt, top=5pt, bottom=5pt,
  toptitle=3pt, bottomtitle=3pt,
  before skip=8pt, after skip=12pt,
  before upper={\raggedright\ttfamily\scriptsize
                \setlength{\parindent}{0pt}\setlength{\parskip}{5pt}},
}

\definecolor{mbparC}{HTML}{356B80}
\definecolor{mbparA}{HTML}{A16B32}
\definecolor{mbparL}{HTML}{785B87}
\definecolor{mbparOff}{HTML}{D2D5D9}
\definecolor{mbcovYes}{HTML}{286449}
\definecolor{mbcovPart}{HTML}{956B25}
\DeclareRobustCommand{\mbAlignedGlyph}[1]{%
  \raisebox{\dimexpr .5ex-.5\height+.5\depth\relax}{#1}}
\DeclareRobustCommand{\mbCameraGlyph}[1]{%
  \mbAlignedGlyph{%
  \tikz[x=0.72em,y=0.72em,baseline=-0.18em,line width=0.45pt,
        draw=#1,line join=round]{%
    \path[use as bounding box] (0,0) rectangle (1.18,1);
    \draw[rounded corners=0.6pt] (0.07,0.12) rectangle (1.1,0.79);
    \draw (0.26,0.79)--(0.36,0.94)--(0.66,0.94)--(0.76,0.79);
    \draw (0.59,0.455) circle (0.22);
    \fill[#1] (0.92,0.65) circle (0.045);
  }}}
\DeclareRobustCommand{\mbActionGlyph}[1]{%
  \mbAlignedGlyph{%
  \tikz[x=1em,y=1em,baseline=-0.18em,line width=0.3pt,
        draw=#1,line join=round]{%
    \path[use as bounding box] (0,0) rectangle (1.5,0.98);
    \foreach \letter/\xx/\yy in {W/0.5/0.5,A/0/0,S/0.5/0,D/1/0}{%
      \draw[rounded corners=0.35pt] (\xx,\yy) rectangle +(0.46,0.46);
      \node[text=#1,inner sep=0pt,font=\sffamily\bfseries\fontsize{3.5}{3.5}\selectfont]
        at (\xx+0.23,\yy+0.23) {\letter};
    }
  }}}
\DeclareRobustCommand{\mbLanguageGlyph}[1]{%
  \mbAlignedGlyph{\textcolor{#1}{\textsf{\textbf{T}}}}}
\newcommand{\mbParadigms}[3]{%
  \ifnum#1=1\relax\mbCameraGlyph{mbparC}\else\mbCameraGlyph{mbparOff}\fi\,
  \ifnum#2=1\relax\mbActionGlyph{mbparA}\else\mbActionGlyph{mbparOff}\fi\,
  \ifnum#3=1\relax\mbLanguageGlyph{mbparL}\else\mbLanguageGlyph{mbparOff}\fi}
\newcommand{\mbHas}{\textcolor{mbcovYes}{$\checkmark$}}
\newcommand{\mbPartial}{\textcolor{mbcovPart}{$\triangle$}}
\newcommand{\mbNot}{\textcolor{black!45}{$\times$}}

\newcommand{\benchname}{WorldMemBench}
\definecolor{mbLinkBackground}{HTML}{F5F6F8}
\definecolor{mbLinkBorder}{HTML}{D5D9E1}
\definecolor{mbLinkText}{HTML}{2C323C}
\newcommand{\mbHuggingFaceIcon}{%
  \begin{tikzpicture}[x=1pt,y=1pt,baseline=-2pt]
    \path[use as bounding box] (0,0) rectangle (15,15);
    \filldraw[fill=yellow!75!orange,draw=orange!80!yellow,line width=.35pt]
      (7.5,8) circle (6.1);
    \fill[mbLinkText] (5.3,9.2) circle (.65) (9.7,9.2) circle (.65);
    \fill[orange!55!red,opacity=.4] (3.5,7) ellipse (1 and .55)
      (11.5,7) ellipse (1 and .55);
    \draw[mbLinkText,line width=.7pt,line cap=round]
      (5.6,6.8) .. controls (6.3,5.2) and (8.7,5.2) .. (9.4,6.8);
    \foreach \xx/\angle in {3/25,12/-25}{%
      \begin{scope}[shift={(\xx,2.8)},rotate=\angle]
        \filldraw[fill=yellow!75!orange,draw=orange!80!yellow,line width=.35pt]
          (0,0) ellipse (2.1 and 1.7);
        \foreach \finger in {-1,0,1}
          \draw[orange!80!yellow,line width=.35pt,line cap=round]
            (\finger,0) -- (\finger,1.2);
      \end{scope}}
  \end{tikzpicture}}
\newcommand{\mbResourceButton}[3]{%
  \href{#1}{\begin{tikzpicture}[baseline=(badge.base)]
    \node[draw=mbLinkBorder,fill=mbLinkBackground,text=mbLinkText,
      line width=.55pt,rounded corners=11pt,minimum height=23pt,
      inner xsep=11pt,inner ysep=3pt,
      font=\bfseries\fontsize{11}{13}\selectfont] (badge)
      {\mbAlignedGlyph{#2}\hspace{5pt}#3};
  \end{tikzpicture}}}
\title{\benchname{}: A Comprehensive Benchmark for Memory of Video World Models}
\author{\textbf{Nan Zhou}$^{1,3}$\thanks{Equal contribution. $^\dagger$Corresponding author.},
\textbf{Jianxiong Gao}$^{2,3}$\footnotemark[1],
\textbf{Qiuyu Wang}$^{3}$,
\textbf{Wen Wang}$^{1}$,
\textbf{Jiahao Wang}$^{4,3}$,\\
\textbf{Yufeng Yuan}$^{4,3}$,
\textbf{Shuaiting Li}$^{1,3}$,
\textbf{Hao Ouyang}$^{3}$,
\textbf{Yujun Shen}$^{3}$,
\textbf{Weiming Lu}$^{1}$,\\
\textbf{Jun Xiao}$^{1}$,
\textbf{Yueting Zhuang}$^{1}$,
\textbf{Yongliang Shen}$^{1\dagger}$
\\[4pt]
$^{1}$Zhejiang University \quad
$^{2}$Fudan University \quad
$^{3}$Ant Group \quad
$^{4}$Nanjing University
}

\iclrfinalcopy % Uncomment for camera-ready version, but NOT for submission.
\begin{document}

\maketitle

\vspace{-0.3in}
\begingroup
\hypersetup{hidelinks}
\centering
\scalebox{0.7}{
    \mbResourceButton{https://zju-real.github.io/WorldMemBench/}{\textcolor{blue!65!cyan}{\faGlobe}}{Homepage}%
    \hspace{30pt}%
    \mbResourceButton{https://github.com/ZJU-REAL/WorldMemBench}{\faGithub}{GitHub}%
    \hspace{30pt}%
    \mbResourceButton{https://huggingface.co/datasets/rescuerz/WorldMemBench}{\mbHuggingFaceIcon}{HuggingFace}
}
\par
\endgroup
% \vspace{0.1in}

\begin{figure}[h]
    \centering
    % \vspace{-0.15in}
    \includegraphics[width=\textwidth]{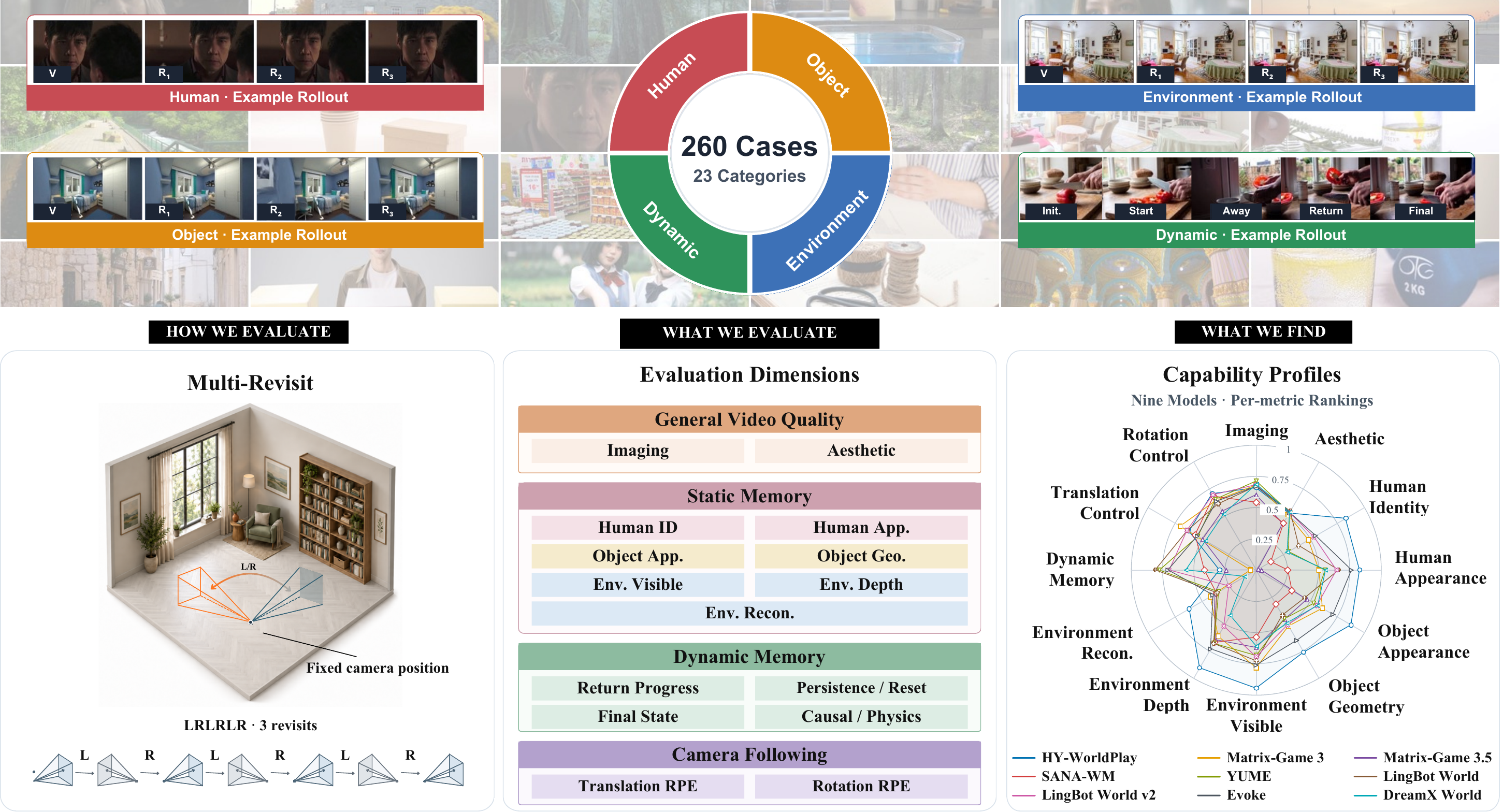}
    \vspace{-0.25in}
    \caption{\textbf{Overview of \benchname}. Top: case coverage across humans, objects,
    environments, and dynamic events. Bottom left: controlled departure-and-revisit
    evaluation compares each return against a fixed initial reference. Bottom right: capability profiles
    of nine video world models.}
    \label{fig:teaser}
    % \vspace{-0.07in}
\end{figure}

\begin{abstract}
\vspace{-0.05in}
As video models enable longer and more realistic interactive experiences, maintaining a world consistent with its history becomes essential. This requires both preserving previously observed content and ensuring coherent world evolution over time. World memory is a critical capability to evaluate alongside visual fidelity and video duration. We introduce \benchname{}, a benchmark for systematically evaluating world memory in video world models. Across 260 test cases, it assesses static memory through the retention of human identity and appearance, object appearance and geometry, and environmental content and geometry, and dynamic memory through event progression and state persistence. Using controlled camera departures and multi-revisit trajectories, the benchmark evaluates static content against fixed initial references and dynamic events against expected state transitions. It combines appearance and geometry metrics with event-specific assessments, and reports video quality and camera-following accuracy as complementary diagnostics. Evaluation of nine representative models shows that strong static retention does not imply reliable dynamic state evolution, while repeated revisits reveal memory degradation beyond the first return. Human studies support the alignment between our automatic evaluations and human judgments. \benchname{} provides a systematic testbed for diagnosing memory failures and guiding the development of persistent, temporally coherent video worlds.
\end{abstract}

\section{Introduction}
Recent advances in autoregressive video generation have enabled longer, more coherent video sequences~\cite{selfforcing,causalforcing,causalforcing++}. In parallel, camera-controlled methods have improved viewpoint and trajectory control~\cite{recammaster}. Together, these developments support video world models that enable sustained, interactive exploration of generated environments~\cite{yume,lingbot-world-v2,Alaya-EVOKE,HYworldplay}. 

Such exploration repeatedly moves content into and out of view, requiring generated worlds to remain consistent with past observations and events. We define \textbf{world memory} as the ability to preserve established content and sustain coherent state evolution across interruptions in observation. Static content, including human identities, object appearance and geometry, and scene layouts, should remain consistent across revisits. Meanwhile, ongoing events should progress coherently while out of view, with their expected consequences reflected upon return. Visual quality and local temporal coherence alone cannot establish these capabilities.

Recent benchmarks have begun to assess memory-related failures,
including revisit consistency~\cite{mbench,r2mbench,worldroambench}
and state evolution under interrupted
observation~\cite{memobench,stevobench,wrbench,liveworld}.
LoopBench~\cite{loopbench} further evaluates scene recall across
multi-revisit trajectories. Building on these efforts, an important question is how memory behaves throughout continued exploration. A successful return demonstrates recovery at one moment, but provides only partial evidence of reliability over subsequent interactions. Repeated revisits help determine whether recovered content remains stable, whether evolving states remain consistent with prior events, and how these behaviors respond to changes in exploration direction. Tab.~\ref{tab:benchmark-comparison} compares benchmark coverage and highlights our joint assessment of memory persistence over extended exploration and repeated revisits, off-screen state evolution, and physical consistency.

To this end, we introduce \textbf{\benchname{}}, a benchmark for systematically evaluating memory in video world models. It comprises 260 cases across 23 semantic categories, including 200 static-memory cases and 60 dynamic-memory cases as in Fig.~\ref{fig:teaser}. Our evaluation covers four dimensions. \textbf{Static Memory} assesses retention of human identity and appearance, object appearance and geometry, and environmental content and geometry. \textbf{Dynamic Memory} evaluates event progression, state persistence, and causal and physical plausibility. \textbf{General Video Quality} and \textbf{Camera Following} serve as complementary diagnostics of perceptual quality and adherence to prescribed exploration trajectories. Reporting these diagnostics separately helps interpret memory performance when rendering quality or camera accuracy differs across models.

In \benchname{}, we use controlled trajectories for leaving and revisiting observed content, including translation loops, repeated rotations about the same axis, and rotations that switch axes between revisits. Each static revisit is compared with the same initial reference to track deviations across successive returns. Dynamic cases specify an initial state, an action, and an expected next state. Case-specific checklists assess event onset and whether subsequent observations reflect its expected progression and consequences. Dynamic memory is thus evaluated through coherent state changes rather than similarity to an unchanged initial frame.

Our evaluation of nine video world models reveals that strong static retention does not necessarily imply reliable dynamic evolution: HY-WorldPlay leads all seven static metrics but initiates only 31.7\% of prescribed events. Repeated revisits expose cumulative degradation that persists on pose-qualified returns, while switching rotation axes adds a further drop of 0.152 in Environment Visible. Human studies show agreement between our automatic metrics and human judgments. We use \benchname{} to examine how context access affects memory, comparing training-free policies while holding the model and generation settings fixed. Combining recent context with view-relevant history improves most evaluated memory metrics and reduces latency relative to full-history access. These results demonstrate the value of \benchname{} for characterizing memory limitations and evaluating context-access strategies.

Our contributions are summarized as follows:
\begin{itemize}[leftmargin=1.2em, labelsep=0.5em]
    \item We introduce \benchname{}, a benchmark comprising 260 cases
    across 23 categories, with 200 static-memory cases covering humans,
    objects, and environments and 60 dynamic-memory cases covering
    seven types of off-screen state transitions.
    \item We design a fixed-reference multi-revisit protocol for static
    memory with same-axis and cross-axis returns, complemented by
    event-specific checklists for dynamic memory. Human evaluation
    yields 78.5\%--92.5\% pairwise agreement across six static metrics
    and Spearman $\rho=0.732$ for the Dynamic Memory Overall score.
    \item We evaluate nine recent video world models and show that no model leads all four dimensions and that memory degrades across successive returns under departure and return controls, while revealing sensitivity to exploration direction and context access.
\end{itemize}

% \input{sections/benchmark_comparison}
% BEGIN inlined sections/benchmark_comparison.tex
% Evidence: analysis/benchmark_comparison_final_checks.md
% BEGIN inlined sections/benchmark_paradigm_icons.tex

% END inlined sections/benchmark_paradigm_icons.tex
\begin{table}
    \centering
    \vspace{-0.1in}
    \caption{\textbf{Comparison with world-model evaluation benchmarks.}
    Input denotes model paradigm:
    \mbCameraGlyph{mbparC} C (camera-driven),
    \mbActionGlyph{mbparA} A (action-driven), and
    \mbLanguageGlyph{mbparL} L (language-driven).
    Gray icons indicate unevaluated paradigms.\mbHas: explicit coverage; \mbPartial: partial coverage; \mbNot: not reported.}
    \label{tab:benchmark-comparison}
    \begingroup
    \fontsize{8}{10}\selectfont
    \setlength{\tabcolsep}{1.5pt}
    \renewcommand{\arraystretch}{1.2}
    \newcommand{\mbTableHead}[1]{\mbAlignedGlyph{\shortstack[c]{#1}}}
    \begin{tabularx}{\textwidth}{@{}>{\raggedright\arraybackslash}X*{8}{c}@{}}
        \toprule
        \mbTableHead{\textbf{Benchmark}}
        & \mbTableHead{\textbf{Input}}
        & \mbTableHead{\textbf{Length}\textbf{(s)}}
        & \mbTableHead{\textbf{Camera}\\\textbf{control}}
        & \mbTableHead{\textbf{Visual}\\\textbf{quality}}
        & \mbTableHead{\textbf{Long-horizon}\\\textbf{revisit}}
        & \mbTableHead{\textbf{State}\\\textbf{evolution}}
        & \mbTableHead{\textbf{Physical}\\\textbf{consistency}}
        & \mbTableHead{\textbf{Multi-}\\\textbf{revisit}} \\
        \midrule
        WorldScore~\cite{worldscore}
        & \mbParadigms{1}{0}{1} & $\sim$2--10
        & \mbHas & \mbHas & \mbNot & \mbNot & \mbNot & \mbNot \\
        MIND~\cite{mind}
        & \mbParadigms{0}{1}{0} & $\sim$5--10
        & \mbHas & \mbHas & \mbPartial & \mbNot & \mbNot & \mbNot \\
        WorldMark~\cite{worldmark}
        & \mbParadigms{1}{1}{1} & 20--60
        & \mbHas & \mbHas & \mbHas & \mbNot & \mbNot & \mbNot \\
        WBench~\cite{wbench}
        & \mbParadigms{1}{1}{1} & $\sim$10--45
        & \mbHas & \mbHas & \mbHas & \mbPartial & \mbPartial & \mbNot \\
        MBench~\cite{mbench}
        & \mbParadigms{1}{1}{1} & $\sim$25
        & \mbHas & \mbNot & \mbHas & \mbHas & \mbHas & \mbNot \\
        MemoBench~\cite{memobench}
        & \mbParadigms{1}{1}{1} & $\sim$4--6
        & \mbHas & \mbHas & \mbNot & \mbHas & \mbHas & \mbNot \\
        WorldRoamBench~\cite{worldroambench}
        & \mbParadigms{1}{1}{1} & 10--60
        & \mbHas & \mbHas & \mbHas & \mbPartial & \mbPartial & \mbNot \\
        WorldExam~\cite{yang2026worldexam}
        & \mbParadigms{1}{1}{1} & $\sim$3--12
        & \mbHas & \mbHas & \mbPartial & \mbPartial & \mbPartial & \mbNot \\
        R2M-Bench~\cite{r2mbench}
        & \mbParadigms{1}{1}{0} & $\sim$30
        & \mbNot & \mbNot & \mbHas & \mbNot & \mbNot & \mbNot \\
        % Source-by-source evidence and unresolved duration fields:
        % analysis/benchmark_additions_20261010.json
        STEVO-Bench~\cite{stevobench}
        & \mbParadigms{1}{0}{1} & $\sim$5--30
        & \mbNot & \mbNot & \mbPartial & \mbHas & \mbHas & \mbNot \\
        WRBench~\cite{wrbench}
        & \mbParadigms{1}{1}{1} & $\sim$4--10
        & \mbHas & \mbNot & \mbPartial & \mbHas & \mbPartial & \mbNot \\
        LiveBench~\cite{liveworld}
        & \mbParadigms{1}{0}{0} & $\sim$16
        & \mbNot & \mbNot & \mbPartial & \mbHas & \mbPartial & \mbHas \\
        LoopBench~\cite{loopbench}
        & \mbParadigms{1}{1}{0} & $\sim$3--24
        & \mbNot & \mbNot & \mbPartial & \mbNot & \mbNot & \mbHas \\
        \midrule
        \textbf{\benchname{}}
        & \mbParadigms{1}{1}{1} & $\sim$20--35
        & \mbHas & \mbHas & \mbHas & \mbHas & \mbHas & \mbHas \\
        \bottomrule
    \end{tabularx}
    \par\vspace{3pt}
    \begin{minipage}{\textwidth}
    \fontsize{8}{9.5}\selectfont
    \end{minipage}
    \endgroup
    \vspace{-0.1in}
\end{table}
% END inlined sections/benchmark_comparison.tex

\section{Related Work}

\subsection{Video World Models}

Self Forcing~\cite{selfforcing} and subsequent advances~\cite{causalforcing,causalforcing++} have improved the efficiency and stability of autoregressive video generation. Together with progress in camera and action control, these developments have made long-horizon interactive generation increasingly practical, enabling a growing class of video world models~\cite{Alaya-EVOKE,HYworldplay,lingbot-world,lingbot-world-v2,matrix-game3,matrix-game35,sanawm}. Beyond generation quality and responsiveness, these models increasingly emphasize spatial memory for maintaining consistency over extended interactions. Relevant approaches include long-context modeling~\cite{lingbot-world}, historical context reconstruction~\cite{HYworldplay}, and field-of-view-based frame retrieval~\cite{contextasmemory}, which help preserve previously observed scene content during exploration.

\subsection{Evaluation of Video World Models}

Video generation is commonly evaluated using FID~\cite{fid} and FVD~\cite{fvd} for distributional similarity, CLIP-based scores~\cite{clipscore} for semantic alignment, and PSNR, SSIM, and LPIPS~\cite{ssim,lpips} for visual fidelity when aligned references are available. Beyond these metrics, benchmarks such as VBench~\cite{vbench}, VBench-2.0~\cite{vbench2}, and WorldScore~\cite{worldscore} provide assessments of visual quality, controllability, text alignment, and physical plausibility. However, performance on these dimensions alone does not establish whether a model can reliably retain and update world information over time. Recent efforts extend evaluation toward world-model capabilities: WBench~\cite{wbench} examines multi-turn interaction and consistency, while WorldMark~\cite{worldmark} standardizes control across heterogeneous model interfaces. WorldModelBench~\cite{worldmodelbench} assesses instruction following and physical plausibility, and WorldExam~\cite{yang2026worldexam} evaluates scene-conditioned reactions and goal-directed behaviors. Recent memory-oriented benchmarks evaluate state evolution under
interrupted observation~\cite{stevobench,wrbench,liveworld} and scene
recall across multi-revisit rollouts~\cite{loopbench}.
MBench~\cite{mbench} and STEVO-Bench~\cite{stevobench} further report that models with strong spatial consistency tend to keep scenes static.
Complementing these efforts, \benchname{} combines fine-grained
Human, Object, and Environment memory with dynamic event evolution,
camera following, and video quality in a unified evaluation framework, and separates event initiation from event continuation when scoring dynamic memory.

% \noindent\textbf{Memory benchmarks.}
% MIND~\cite{mind} evaluates scene consistency through symmetric traversal and
% action-space generalization. MBench~\cite{mbench} uses trigger-conditioned
% scoring for entity, environment, and causal consistency, while MemoBench~\cite{memobench} evaluates
% object reappearance and evolving physical states through a
% disappear-and-reappear protocol. R2M-Bench~\cite{r2mbench} calibrates revisit
% consistency against temporal controls from the same rollout; WorldRoamBench~\cite{worldroambench}
% accounts for imperfect action execution through trajectory-aware scene and
% subject memory evaluation. 
% Tab.~\ref{tab:benchmark-comparison} compares their model paradigms, evaluated video lengths, and capability coverage. \benchname{} combines entity and environment consistency, dynamic memory, visual quality, and camera following with a fixed-reference multi-revisit protocol to measure retention at each successive return across human, object, and environment targets, compares same-axis and cross-axis revisits, and separately evaluates event progression during absence. Video quality and camera following provide complementary diagnostics.

\section{\benchname{} Construction}

In this section, we describe the construction of \benchname{}, including its data composition and camera trajectory design. The benchmark evaluates world memory through controlled departure-and-revisit rollouts, probing whether previously observed content and evolving states remain consistent across periods outside the camera's field of view.

\subsection{Evaluation Dimensions}
\label{sec:Evaluation-dimension}

Inspired by hierarchical diagnostic evaluation~\cite{yang2026worldexam}, we organize evaluation into four complementary dimensions. \textbf{Static Memory} assesses the retention of previously observed humans, objects, and environments upon revisitation, covering individual entities through global scene structure. \textbf{Dynamic Memory} examines whether events progress appropriately while out of view and whether their consequences persist upon return. Each case specifies an initial state, an action, and a next state, with assessment covering event progression, state persistence, final-state correctness, and causal and physical consistency. These two memory dimensions are complemented by \textbf{General Video Quality}, which measures imaging and aesthetic quality, and \textbf{Camera Following}, which evaluates adherence to the prescribed camera motion. Together, they characterize memory performance alongside the visual quality and control accuracy of the generated rollouts.

\begin{figure}
% \vspace{-0.1in}
    \centering
    \includegraphics[width=\textwidth]{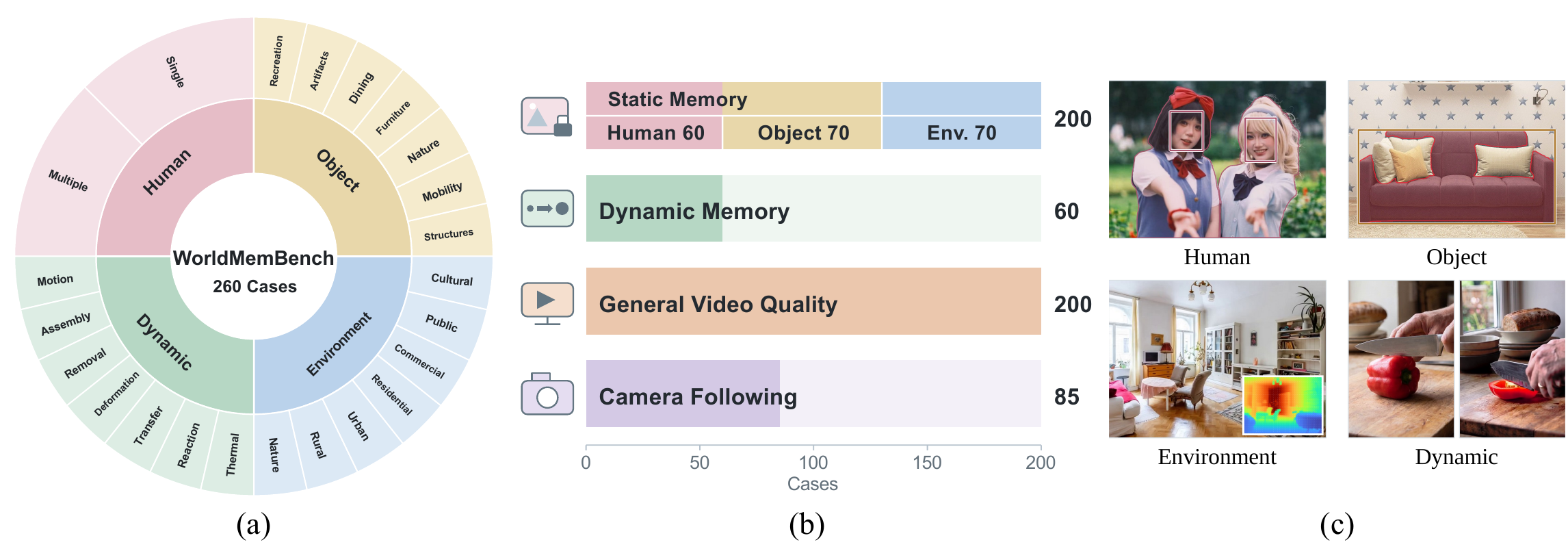}
    \vspace{-0.15in}
    \caption{
        \textbf{Composition and diversity of \benchname{}.}
        \textbf{(a)} Case taxonomy, with Human, Object, Environment, and Dynamic categories in the inner ring and subcategories in the outer ring.
        \textbf{(b)} Case counts for static memory, dynamic memory, general video quality, and camera-following evaluation.
        \textbf{(c)} Representative cases showing target boxes and masks, scene depth, and event states before and after revisiting.
    }
    \label{fig:benchmark-statistics}
    \vspace{-0.1in}
\end{figure}

\begin{figure}
    \centering
    \includegraphics[width=\textwidth]{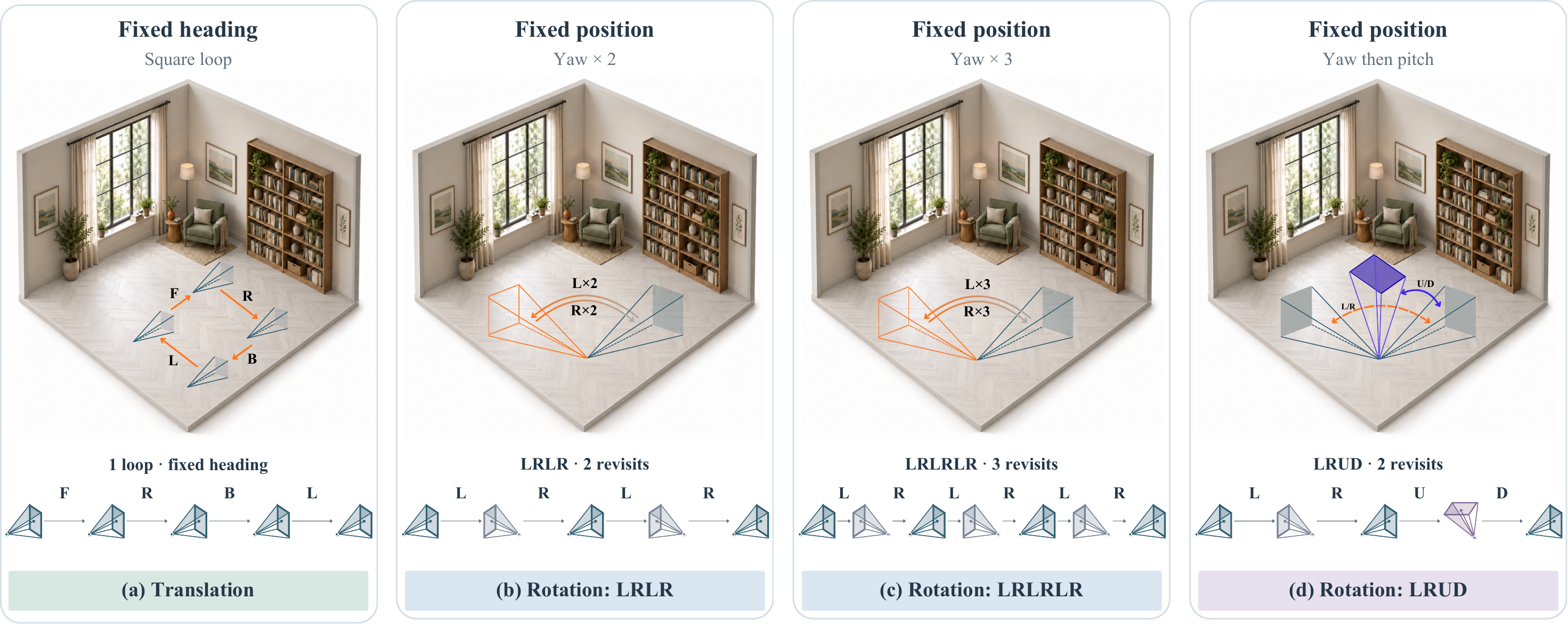}
    \vspace{-0.15in}
    \caption{\textbf{Evaluation camera trajectories in \benchname{}}.
    Translation preserves the initial viewing direction and completes a square loop (\textcolor{teal}{green}), while Rotation keeps the camera position fixed and applies left--right yaw (\textcolor{orange}{orange}) or up--down pitch (\textcolor{blue}{blue}). 
    }
    \label{fig:camera-trajectories}
    \vspace{-0.10in}
\end{figure}

\subsection{Data Composition}
\label{sec:Data-Composition}

\benchname{} comprises 260 core cases across four families: 60 Human, 70 Object, 70 Environment, and 60 Dynamic Memory cases. The first three families form the 200-case Static Memory subset. General Video Quality is evaluated on all static-memory rollouts, while Camera Following reuses 85 static-memory cases suitable for trajectory evaluation. As shown in Fig.~\ref{fig:benchmark-statistics}, the four case families jointly cover 23 representative content categories. Human cases distinguish single- and multi-person scenes; Object and Environment cases span diverse target entities and spatial settings; and Dynamic Memory covers seven state-transition mechanisms. These 23 categories provide coverage across entity types, scene configurations, and dynamic processes.

\subsection{Camera Trajectories}
\label{sec:Camera-Trajectories}
We construct translation and rotation trajectories to move the camera away from a previously observed region and subsequently return to it, as illustrated in Fig.~\ref{fig:camera-trajectories}. We denote the initial observation phase by $V$, the $i$-th departure by $D_i$, and the corresponding prescribed revisit by $R_i$, providing shared temporal anchors for memory evaluation. \textbf{Translation} maintains a fixed camera orientation and follows a Forward--Right--Backward--Left sequence, forming a square loop that returns to the initial position at $R_1$. \textbf{Rotation} keeps the camera position fixed while changing its viewing direction. Here, \textbf{L}/\textbf{R} denote left/right yaw rotations, and \textbf{U}/\textbf{D} denote upward/downward pitch rotations. Each \textbf{LR} or \textbf{UD} pair turns the camera away from and then back to its initial viewing direction, completing one departure-and-revisit cycle. Accordingly, \textbf{LRLR} and \textbf{LRLRLR} repeat the left--right cycle twice and three times, respectively. \textbf{LRUD} performs a left--right cycle followed by an up--down cycle, yielding two revisits across different rotation axes.

\section{Evaluation Protocol and Metrics}

We describe the evaluation protocol and metrics for video quality, static memory, dynamic memory, and camera following.

\subsection{Multi-Revisit Protocol}
\label{sec:multi-revisit-protocol}

% For each generated rollout, the prescribed camera trajectory defines $K$ revisits.
% Let $\mathcal{U}$ denote its evaluation targets. For target $u$, the initial
% visible observations $V_u$ form a frozen reference, and $R_{u,k}$ denotes the
% evidence associated with the $k$-th revisit. Each metric defines a
% revisit-level consistency score $s_{u,k}$ from these two sets.We aggregate uniformly over revisits and targets:
% \begin{equation}
%     S_u
%     = \frac{1}{K}\sum_{k=1}^{K}s_{u,k},
%     \qquad
%     S_{\mathrm{rollout}}
%     = \frac{1}{|\mathcal{U}|}
%       \sum_{u\in\mathcal{U}}S_u.
%     \label{eq:revisit-macro}
% \end{equation}
% The aggregate summarizes performance across revisits, while the individual scores characterize retention at each return. Here, $K$ counts all prescribed revisits, including those without valid return observations. 

For each rollout, we evaluate the revisits specified by the camera trajectory.
At each revisit, the observed content is compared with a fixed reference from the initial observation phase.
We compute scores for each target to track retention across returns.
The overall score is obtained by averaging over revisits and then over targets:
\begin{equation}
    S_u = \frac{1}{K}\sum_{k=1}^{K}s_{u,k},
    \qquad
    S_{\mathrm{rollout}}
    = \frac{1}{|\mathcal{U}|}
      \sum_{u\in\mathcal{U}}S_u,
    \label{eq:revisit-macro}
\end{equation}
where $\mathcal{U}$ is the set of targets, $K$ is the number of prescribed revisits, and $s_{u,k}$ is the consistency score for target $u$ at the $k$-th revisit.
$S_u$ is the average score for target $u$, and $S_{\mathrm{rollout}}$ is the overall rollout score.
All prescribed revisits are included in $K$, including those without valid observations.

\subsection{General Video Quality}
\label{sec:metric-general-quality}

We assess visual quality by sampling every fourth frame of each rollout. \textbf{Imaging Quality} is measured using MUSIQ-SPAQ~\cite{ke2021musiqmultiscaleimagequality}, while \textbf{Aesthetic Quality} uses the LAION aesthetic predictor based on CLIP features~\cite{radford2021clip}. Scores are averaged over sampled frames within each video, then equally across cases. Both metrics are normalized to $[0,1]$ and reported separately, with higher values indicating better quality.

\subsection{Camera Following}
\label{sec:metric-camera-following}

Camera Following measures adherence to prescribed camera motion, supporting interactive exploration and helping identify viewpoint errors that can confound memory evaluation. We report translation and rotation relative pose errors (RPE).
Let $Q_t$ and $\widehat Q_t^{(e)}$ denote the target and estimated
camera-to-world poses at frame $t$, where $e$ identifies the pose estimator.

\noindent\textbf{1) Translation}, we normalize the target path to unit length and align
the DA3~\cite{depthanything3} and ViPE~\cite{vipe} trajectories independently
using $\mathrm{Sim}(3)$, removing differences in global coordinate frame and
scale. Let $\Delta Q_t=Q_t^{-1}Q_{t+1}$ and
$\Delta\widehat Q_t^{(e)}=(\widehat Q_t^{(e)})^{-1}
\widehat Q_{t+1}^{(e)}$ denote adjacent target and estimated motions. Their
translation disagreement over the full rollout is:
\begin{equation}
    \mathrm{RPE}_{\mathrm{trans}}
    = \frac{1}{2}\sum_{e\in\{\mathrm{DA3},\mathrm{ViPE}\}}
      \sum_{t=0}^{T-2}
      \left\|\operatorname{trans}\!\left(
      (\Delta Q_t)^{-1}\Delta\widehat Q_t^{(e)}
      \right)\right\|_2.
    \label{eq:translation-rpe-main}
\end{equation}
\noindent\textbf{2) Rotation}, only the DA3 orientations are used, without
$\mathrm{Sim}(3)$ alignment. Let $R_t$ and $\widehat R_t$ be the target and
estimated rotations, with adjacent increments
$\Delta R_t=R_t^{\top}R_{t+1}$ and
$\Delta\widehat R_t=\widehat R_t^{\top}\widehat R_{t+1}$. We normalize their
accumulated angular disagreement by the target angular path:
\begin{equation}
    \mathrm{RPE}_{\mathrm{rot}}
    = \frac{\sum_{t=0}^{T-2}
      \theta\!\left((\Delta R_t)^{\top}\Delta\widehat R_t\right)}
      {\sum_{t=0}^{T-2}\theta(\Delta R_t)}.
    \label{eq:rotation-rpe-main}
\end{equation}
Here, $T$ is the number of frames, $\operatorname{trans}(\cdot)$ extracts the
translation component, and $\theta(\cdot)$ is the $\mathrm{SO}(3)$ geodesic
angle. Both metrics are dimensionless and lower is better.

\subsection{Static Memory}
\label{sec:metric-static-memory}

Static Memory evaluates the retention of previously observed world content across departures and revisits. All metrics use fixed references established during the initial observation phase, score each prescribed revisit independently, and aggregate the results using Eq.~\ref{eq:revisit-macro}.

\noindent\textbf{Human Memory.} We evaluate each person identified in the conditioning frame.
\textbf{1) Identity} uses ArcFace~\cite{arcface} face embeddings.
\textbf{2) Appearance} uses DINOv2~\cite{dinov2} features extracted within person masks to assess clothing and overall appearance.
Both metrics compare valid revisit observations with initial references:
\begin{equation}
    s_k
    = \frac{1}{|R_k|}\sum_{t\in R_k}
      \operatorname{clip}_{[0,1]}\!\left(
      \max_{r\in V}
      \frac{\phi(I_r)^{\top}\phi(I_t)}
      {\|\phi(I_r)\|_2\,\|\phi(I_t)\|_2}
      \right),
    \label{eq:human-revisit-score}
\end{equation}
where $I_t$ is a valid observation of the person at frame $t$, $\phi$ is the corresponding encoder, and $V$ and $R_k$ denote the initial reference set and the $k$-th revisit set, respectively.
The reference set $V$ remains fixed throughout the rollout, and revisits without valid observations receive a score of zero.
Identity and Appearance are aggregated separately using Eq.~\ref{eq:revisit-macro}, with equal weight for each person.
Both metrics lie in $[0,1]$, with higher scores indicating better retention.

\noindent\textbf{Object Memory.} The target object is specified by a mask in the conditioning frame and located at each revisit using semantic object detection and video segmentation.
\textbf{1) Appearance} follows the fixed-reference comparison in Eq.~\ref{eq:human-revisit-score}, with $\phi$ extracting DINOv2 features within object masks.
\textbf{2) Geometry Consistency} compares object silhouettes after cropping each mask to its tight bounding box, resizing it while preserving its aspect ratio, and centering it on a common canvas:
\begin{equation}
    s_k^{\mathrm{geo}}
    = \frac{1}{|R_k|}\sum_{t\in R_k}
      \max_{r\in V}
      \operatorname{IoU}\!\left(\mathcal C(M_r),\mathcal C(M_t)\right),
    \label{eq:object-geometry}
\end{equation}
where $M_t$ is the object mask at frame $t$ and $\mathcal C(\cdot)$ denotes the normalization described above.
This normalization removes image-plane translation and uniform scale differences while preserving silhouette changes.
Both metrics lie in $[0,1]$, with higher scores indicating better retention.

\noindent\textbf{Environment Memory.}
We evaluate scene retention through appearance, depth ordering, and 3D
structure. 
\textbf{1) Visible Memory} compares global DINOv2 features between the
conditioning frame and each return. Within each prescribed rotational return
window, we use the DA3 camera poses to select the frame whose estimated camera
orientation is closest to that of the conditioning frame. Visible Memory is the
mean of the clipped cosine similarities between the selected return frames and
the conditioning frame.
\textbf{2) Depth Spearman} computes the rank correlation between depth values at
corresponding image locations in the initial and revisit views, measuring preservation of
relative depth ordering independently of absolute scale.
\textbf{3) Spherical Reconstruction Score} projects reference and revisit depth observations onto a shared spherical grid and compares reconstructed surfaces after global scale alignment, penalizing missing surfaces, additional surfaces, and distance discrepancies. Each angular bin stores the nearest log-range. Let $H$ and $G_k$ denote the reference
and $k$-th revisit fields, while $\mathcal C_k$ and $\mathcal U_k$ denote the
intersection and union of their occupied bins. We remove the global depth scale
with
$\delta_k=\operatorname{median}_{b\in\mathcal C_k}(H_b-G_{k,b})$. The three metrics are reported separately.
Higher values indicate better retention for all three metrics.
\begin{equation}
\begin{aligned}
    S_{\mathrm{depth}}
    &=\frac{1}{K}\sum_{k=1}^{K}
      \rho_{\mathrm S}\!\left(d_0,d_{r_k}\right),
    \qquad
    S_{\mathrm{rec}}
    &=\frac{1}{K}\sum_{k=1}^{K}
      \frac{|\mathcal C_k|}{|\mathcal U_k|}
      \operatorname*{mean}_{b\in\mathcal C_k}
      \exp\!\left(-\left|H_b-(G_{k,b}+\delta_k)\right|\right)
      .
\end{aligned}
\label{eq:environment-geometry}
\end{equation}

\FloatBarrier
\begin{table*}[!t]
\vspace{-0.1in}
\centering
\caption{
Quantitative comparison of video world models on \benchname{}. $\uparrow$ / $\downarrow$ indicate higher / lower is better. The best and second-best results for each metric are shown in \textbf{bold} and \underline{underlined}, respectively. App., Env., Recon., Trans., Rot., Prog., Persist., and Phys. denote Appearance, Environment, Reconstruction, Translation, Rotation, Progress, Persistence, and Physics, respectively.
}
\label{tab:main-results}

\begingroup
\setlength{\aboverulesep}{0pt}
\setlength{\belowrulesep}{0pt}

\resizebox{\textwidth}{!}{%
\begin{tabular}{l|*{9}{c}}
\toprule[0.9pt]
\addlinespace[5pt]

\multirow{2}{*}{\textbf{Metric}}
& \textbf{HY}
& \textbf{Matrix}
& \textbf{Matrix}
& \multirow{2}{*}{\textbf{SANA-WM}}
& \multirow{2}{*}{\textbf{YUME}}
& \textbf{LingBot}
& \textbf{LingBot}
& \multirow{2}{*}{\textbf{Evoke}}
& \textbf{DreamX} \\

& \textbf{WorldPlay}
& \textbf{Game-3}
& \textbf{Game-3.5}
&
&
& \textbf{World}
& \textbf{World-v2}
&
& \textbf{World} \\

\addlinespace[5pt]
\specialrule{0.5pt}{0pt}{3pt}

% ================= General Video Quality =================
\multicolumn{10}{l}{%
  \rule{0pt}{12.5pt}\textcolor[HTML]{477F85}{\rule{2pt}{8pt}\hspace{5pt}\textbf{Video Quality}}} \\

Imaging $\uparrow$
& 0.659 & 0.680 & 0.603 & 0.542 & \cellcolor[HTML]{E7F3F4}\textbf{0.713}
& 0.671 & 0.683 & 0.670 & \underline{0.686} \\

Aesthetic $\uparrow$
& 0.527 & 0.510 & 0.429 & 0.434 & 0.533
& \cellcolor[HTML]{E7F3F4}\textbf{0.545} & 0.534 & 0.533 & \underline{0.537} \\

\specialrule{0.45pt}{3pt}{3pt}

% ===================== Static Memory =====================
\multicolumn{10}{l}{%
  \rule{0pt}{12.5pt}\textcolor[HTML]{80639B}{\rule{2pt}{8pt}\hspace{5pt}\textbf{Static Memory}}} \\

Human ID $\uparrow$
& \cellcolor[HTML]{EDE5F7}\textbf{0.828} & 0.483 & 0.019 & 0.134 & 0.298
& 0.393 & 0.534 & \underline{0.550} & 0.286 \\

Human App. $\uparrow$
& \cellcolor[HTML]{EDE5F7}\textbf{0.830} & 0.502 & 0.043 & 0.251 & 0.551
& 0.661 & 0.645 & \underline{0.762} & 0.556 \\

Object App. $\uparrow$
& \cellcolor[HTML]{EDE5F7}\textbf{0.879} & 0.612 & 0.471 & 0.328 & 0.531
& 0.448 & 0.564 & \underline{0.710} & 0.569 \\

Object Geometry $\uparrow$
& \cellcolor[HTML]{EDE5F7}\textbf{0.761} & 0.518 & 0.457 & 0.314 & 0.449
& 0.418 & 0.506 & \underline{0.651} & 0.491 \\

Env. Visible $\uparrow$
& \cellcolor[HTML]{EDE5F7}\textbf{0.947} & \underline{0.782} & 0.620 & 0.537 & 0.682
& 0.728 & 0.694 & 0.762 & 0.610 \\

Env. Depth $\uparrow$
& \cellcolor[HTML]{EDE5F7}\textbf{0.905} & 0.611 & 0.645 & 0.668 & 0.676
& 0.680 & 0.517 & \underline{0.730} & 0.419 \\

Env. Recon. $\uparrow$
& \cellcolor[HTML]{EDE5F7}\textbf{0.619} & \underline{0.423} & 0.390 & 0.367 & 0.347
& 0.364 & 0.253 & 0.400 & 0.105 \\

\specialrule{0.45pt}{3pt}{3pt}

% ==================== Dynamic Memory =====================
\multicolumn{10}{l}{%
  \rule{0pt}{12.5pt}\textcolor[HTML]{A38128}{\rule{2pt}{8pt}\hspace{5pt}\textbf{Dynamic Memory}}} \\

Overall $\uparrow$
& 0.294 & 0.046 & 0.240 & 0.415 & \underline{0.775}
& \cellcolor[HTML]{FFF3CC}\textbf{0.808} & 0.696 & 0.706 & 0.563 \\

Return Prog. $\uparrow$
& 0.292 & 0.050 & 0.183 & 0.342 & \underline{0.708}
& \cellcolor[HTML]{FFF3CC}\textbf{0.775} & 0.667 & 0.667 & 0.558 \\

Persist./Reset $\uparrow$
& 0.308 & 0.050 & 0.233 & 0.458 & \underline{0.817}
& \cellcolor[HTML]{FFF3CC}\textbf{0.833} & 0.733 & 0.717 & 0.600 \\

Final State $\uparrow$
& 0.267 & 0.042 & 0.200 & 0.333 & \underline{0.717}
& \cellcolor[HTML]{FFF3CC}\textbf{0.750} & 0.592 & 0.658 & 0.492 \\

Causal/Phys. $\uparrow$
& 0.308 & 0.042 & 0.342 & 0.525 & \underline{0.858}
& \cellcolor[HTML]{FFF3CC}\textbf{0.875} & 0.792 & 0.783 & 0.600 \\

\specialrule{0.45pt}{3pt}{3pt}

% ===================== Camera Follow =====================
\multicolumn{10}{l}{%
  \rule{0pt}{12.5pt}\textcolor[HTML]{527BAB}{\rule{2pt}{8pt}\hspace{5pt}\textbf{Camera Follow}}} \\

Trans. RPE $\downarrow$
& 0.609 & \cellcolor[HTML]{E2EDFC}\textbf{0.424} & 1.018 & 0.771 & 0.853
& 0.575 & \underline{0.565} & 0.838 & 1.139 \\

Rot. RPE $\downarrow$
& \cellcolor[HTML]{E2EDFC}\textbf{0.417} & 0.554 & 0.837 & 0.464 & 0.529
& 0.637 & \underline{0.439} & 0.573 & 0.928 \\

\specialrule{0.9pt}{3pt}{0pt}
\end{tabular}%
}
\endgroup
% \vspace{-0.1in}
\end{table*}

\subsection{Dynamic Memory}
\label{sec:metric-dynamic-memory}
Dynamic Memory evaluates whether an event continues to evolve while out of view and whether its consequences remain visible after return. Each case specifies an initial state, an action, and an expected next state. We use Qwen3.8-27B~\cite{qwen38} to generate candidate binary questions and select eight questions with fixed reference answers, covering four aspects with two questions each: return progress, persistence/reset, final state, and causal/physical consistency. During evaluation, Qwen3.8-27B first determines whether the target event visibly starts during the initial hold and then answers all eight questions from the ordered rollout evidence. The event-start decision acts as a gate: if the event does not start, all eight question scores are zero; otherwise, each answer receives one point if it matches the fixed reference answer and zero otherwise. Each sub-dimension score is the mean of the two gated scores for that aspect, while Overall is the mean of all eight gated scores. All scores lie in $[0,1]$, with higher values indicating better performance. The complete prompts and event-start decomposition are provided in Appendix~\ref{app:dynamic-memory-prompts}.

% Dynamic Memory 评测不可见期间发生的状态演化。每个 case 包含 initial state、action
% 和 next state，并配有八个冻结的二元问题，分别覆盖 return progress、
% persistence/reset、final state 与 causal/physics 四个方面。回答与冻结答案一致记为
% 1，否则记为 0；四个方面各由两题均值
% 得到，Overall 为全部八题的等权平均。所有 Dynamic Memory 分数均在 $[0,1]$，
% 越高越好。

% 对于 Dynamic Memory，我们使用 Qwen3.8-27B 为每个 case 生成候选问题，从中筛选并
% 冻结八个二元问题，覆盖 return progress、persistence/reset、final state 和
% causal/physical consistency；正式评测仍由 Qwen3.8-27B 依据 rollout 中的可见证据
% 逐题作答，最终分数为预测答案与冻结答案的一致率。完整 prompt 见
% Appendix~\ref{app:dynamic-memory-prompts}。

% \newpage
\section{Experiment}
% In this section, we evaluate world models using our proposed \benchname.

\subsection{Setup}
We evaluate nine models across all dimensions of \benchname{}:
HY-WorldPlay~\cite{HYworldplay},
Matrix-Game 3~\cite{matrix-game3},
Matrix-Game 3.5~\cite{matrix-game35},
SANA-WM~\cite{sanawm},
YUME~\cite{yume},
LingBot-World~\cite{lingbot-world},
LingBot-World-v2 \\ ~\cite{lingbot-world-v2},
Evoke~\cite{Alaya-EVOKE}, and
DreamX-World~\cite{DreamX-World}. 
All models are evaluated under consistent benchmark conditions using a unified evaluation protocol, with each model configured
according to the defaults provided in its official release. Model-specific resolutions, frame rates, temporal generation strategies, control adaptations, and inference settings are detailed in Appendix~\ref{app:inference-settings}.

\subsection{Evaluation of World Models}
\label{sec:Evaluation}

\noindent \textbf{Quantitative Results.}
Following the evaluation protocol, we evaluate videos generated by nine world models across four dimensions: General Video Quality, Static Memory, Dynamic Memory, and Camera Following. 
Tab.~\ref{tab:main-results} reveals distinct strengths across these dimensions. HY-WorldPlay leads all seven Static Memory metrics, demonstrating strong retention of human, object, and environmental content. However, its Dynamic Memory score is substantially lower than those of LingBot-World and YUME, which achieve the highest and second-highest scores, respectively. These results suggest that
preserving static content and maintaining evolving states pose distinct challenges. Camera-following performance also varies: Matrix-Game 3 achieves the lowest translation error, while HY-WorldPlay achieves the lowest rotation error. Overall, no model leads across all four dimensions, highlighting the need to evaluate complementary capabilities separately.

\noindent \textbf{Qualitative Results.}
To examine whether our metrics reflect observable memory behavior, we present an Environment multi-revisit comparison in Fig.~\ref{fig:vis_3}, with Human and Object comparisons in Appendix~\ref{app:qualitative-results}. In the Human, Object, and Environment cases shown in Figs.~\ref{fig:vis_1}, \ref{fig:vis_2}, and \ref{fig:vis_3}, HY-WorldPlay most consistently preserves subject identity, object geometry, and scene layout across three revisits. These observations align with its first-place performance on all seven Static Memory metrics in Tab.~\ref{tab:main-results}, providing qualitative support for our quantitative evaluation of memory consistency across different content types and repeated revisits.

\begin{figure}[!t]
    \vspace{-0.10in}
    \centering
    \includegraphics[width=\linewidth,trim={8pt 75pt 12pt 10pt},clip]{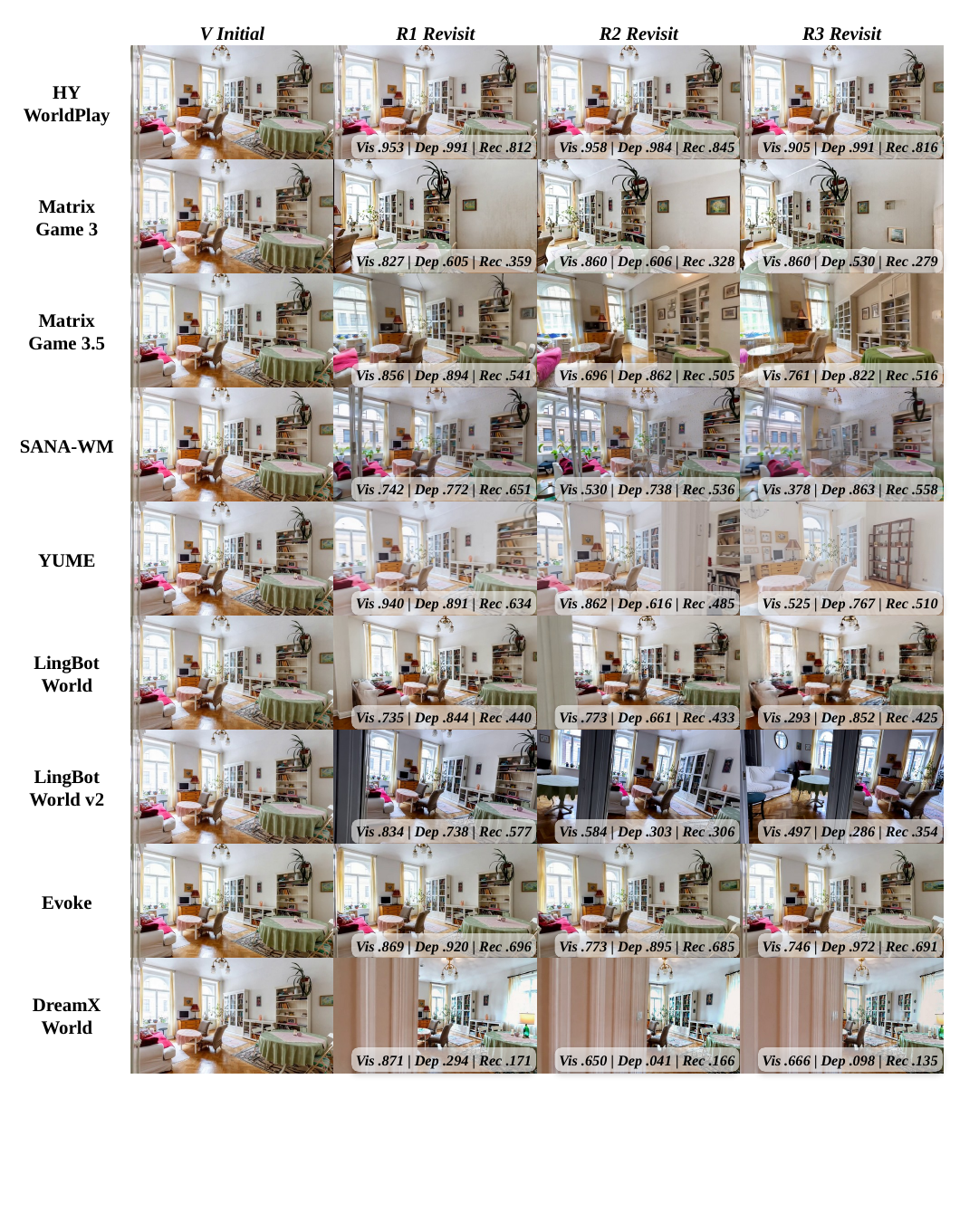}
    \vspace{-0.25in}
    \caption{
        Qualitative comparison of video world models on an environment case.
    }
    \label{fig:vis_3}
    % \vspace{-0.1in}
\end{figure}

\noindent \textbf{Human Evaluation.}
To complement the qualitative analysis, we assess how well our Static and Dynamic Memory metrics align with human judgments. For Static Memory (Tab.~\ref{tab:static-human-alignment}), blinded A/B agreement ranges from 78.5\% to 92.5\% across the six metrics, with model-level Spearman correlations ranging from $0.301$ (Environment Depth) to $0.988$ (Environment Visible).
% of $0.988$, $0.952$, and $0.929$ for Environment Visible, Human Appearance, and Human Identity, respectively. 
For Dynamic Memory (Tab.~\ref{tab:dynamic-human-alignment}), the checklist judge achieves a Spearman correlation of $0.732$ and a Pearson linear correlation coefficient (PLCC) of $0.808$ for the Overall score across 50 instances. Complete annotation protocols, per-dimension results, and interfaces are provided in Appendix~\ref{app:human-eval-interface}. Together, these results support the validity of our metrics for assessing world memory.

\subsection{Multi-Revisit Analysis}
\label{sec:multi-revisit-analysis}

\begin{figure}[t]
    \centering
    \includegraphics[width=\textwidth]{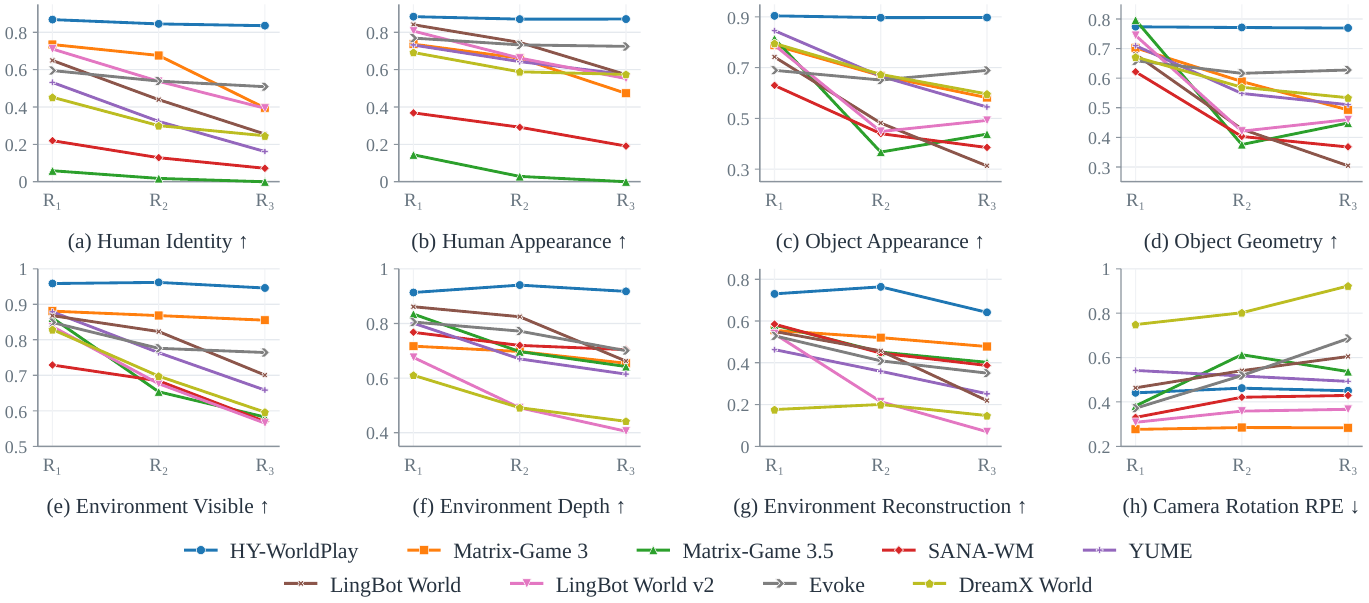}
    \vspace{-0.25in}
    \caption{Per-model changes of nine models across the three revisits of LRLRLR, four metrics per row. $\uparrow$/$\downarrow$ indicate higher/lower is better; all panels keep each metric’s original value direction.}
    \label{fig:repeated-revisit-lines}
\end{figure}

Fig.~\ref{fig:repeated-revisit-lines} shows how the seven Static Memory
metrics and Camera Rotation RPE change across nine models from $R_1$ to
$R_3$ under LRLRLR, revealing two main findings: 
\textbf{1) Multi-Revisit Evaluation Exposes Cumulative Memory Decay.}
Nine-model averages degrade monotonically across revisits on all eight metrics: Object Appearance, Human Identity, and Object Geometry drop by 0.229, 0.217, and 0.205 from $R_1$ to $R_3$, while Environment Reconstruction Score decreases by 0.196. The trend also holds per model: on the Human, Object, Environment Visible, and Environment Reconstruction metrics, every model is weaker at $R_3$ than at $R_1$, and Environment Depth degrades for eight of the nine models. Content recovered on the first return therefore does not necessarily persist across later revisits; multi-revisit evaluation exposes cumulative memory decay that single-return evaluation cannot observe.
\textbf{2) Cumulative decay stems from compounding viewpoint misalignment and autoregressive error propagation.}
First, the nine-model mean Camera Rotation RPE grows from 0.430 to 0.502 and 0.530, so consecutive rotations make the returning viewpoint progressively harder to align with the initially observed region. Second, all nine models generate long videos autoregressively: each new segment conditions on previously generated content, so early appearance, structure, and geometry errors propagate into later conditioning and impose longer error-propagation chains on later revisits. Even among the 431 Environment model--case pairs satisfying the departure
and return criteria, the nine-model mean Visible, Depth, and Reconstruction
scores decrease by 0.152, 0.135, and 0.139 from $R_1$ to $R_3$, respectively
(Appendix~\ref{app:revisit-controls}).

\subsection{Cross-Axis Revisit Robustness}
\label{sec:cross-axis-revisit-analysis}

LRLR and LRUD share the first yaw revisit but differ at the second: LRLR repeats
the yaw cycle, whereas LRUD returns through pitch. For each higher-is-better
Static Memory metric, we define $d=R_1-R_2$ and
$\Delta_{\mathrm{axis}}=d_{\mathrm{LRUD}}-d_{\mathrm{LRLR}}$. Across nine
models, mean $\Delta_{\mathrm{axis}}$ is positive for all seven metrics, and 50
of 63 model--metric pairs degrade more under LRUD; Environment Visible shows the
largest additional degradation at 0.152. Switching exploration axes therefore
introduces difficulty beyond same-axis multi-revisit evaluation. Trajectory-wise means
and full per-model results are provided in Appendix~\ref{app:trajectory-results}.Appendix~\ref{app:revisit-controls} reports departure/return coverage and evaluates both trends on tracking-supported and pose-qualified subsets.

\subsection{Diagnostic Case Study: Training-Free Context Access}
\label{sec:lingbot-v2-context-management}
The main results reveal distinct performance profiles for static content recovery and dynamic state evolution. Budgeted retrieval of view-relevant history has been proposed to extend the memory of video world models~\cite{contextasmemory,yi2026worldkvefficientworldmemory}; we use \benchname{} to examine which memory dimensions such policies help. To examine how context access affects these capabilities, we compare three training-free policies on LingBot World v2, keeping the model and other generation settings fixed. FullKV provides access to the complete history. Sliding Windows uses recent context within a fixed budget, while FoV combines recent context with view-relevant historical chunks under the same budget. FoV achieves the best scores on eight of the eleven capability metrics and outperforms Sliding Windows on all eleven. Compared with FullKV, it scores higher on all four Dynamic Memory sub-metrics and both Human metrics. FullKV, however, retains an advantage in Object Appearance, Object Geometry, and Environment Visible, suggesting that the benefits of context selection vary across memory dimensions. Both budgeted policies also reduce inference time. Across 660 matched videos, mean end-to-end latency is 322.2\,s for FullKV, compared with 262.7\,s for FoV and 260.4\,s for Sliding Windows. These correspond to speedups of $1.23\times$ and $1.24\times$, respectively. Overall, the results suggest that combining recent context with view-relevant history offers a favorable balance between memory performance and efficiency. Context budgets, evaluation sets, timing protocols, and further analysis are provided in Appendix~\ref{app:context-policy-analysis}.

\begin{table}
\vspace{-0.15in}
    \centering
    \caption{Controlled comparison of context-access policies on LingBot World v2. Best and second-best results are \textbf{bold} and \underline{underlined}.}
    \label{tab:lingbot-v2-context-management}
    \scriptsize
    \setlength{\tabcolsep}{2.2pt}
    \renewcommand{\arraystretch}{1.05}
    \resizebox{\textwidth}{!}{%
    \begin{tabular}{@{}l*{13}{c}@{}}
        \toprule
        \multirow{2}{*}{Config}
            & \multicolumn{2}{c}{Human}
            & \multicolumn{2}{c}{Object}
            & \multicolumn{3}{c}{Environment}
            & \multicolumn{4}{c}{Dynamic Memory}
            & \multicolumn{2}{c}{Inference Speed} \\
        \cmidrule(lr){2-3}\cmidrule(lr){4-5}\cmidrule(lr){6-8}
        \cmidrule(lr){9-12}\cmidrule(lr){13-14}
            & ID $\uparrow$ & App. $\uparrow$
            & App. $\uparrow$ & Geometry $\uparrow$
            & Visible $\uparrow$ & Depth $\uparrow$ & Recon. $\uparrow$
            & Return $\uparrow$ & Persist. $\uparrow$ & Final $\uparrow$ & Causal $\uparrow$
            & Latency (s) $\downarrow$ & Speedup $\uparrow$ \\
        \midrule
        FullKV
            & \underline{0.534} & \underline{0.645}
            & \textbf{0.564} & \textbf{0.506}
            & \textbf{0.694} & \underline{0.517} & \underline{0.253}
            & \underline{0.667} & \underline{0.733} & 0.592 & \underline{0.792}
            & 322.16 & 1.00$\times$ \\
        FoV
            & \textbf{0.547} & \textbf{0.663}
            & \underline{0.531} & \underline{0.485}
            & \underline{0.687} & \textbf{0.550} & \textbf{0.287}
            & \textbf{0.725} & \textbf{0.792} & \textbf{0.675} & \textbf{0.800}
            & \underline{262.69} & \underline{1.23$\times$} \\
        Sliding Windows
            & 0.392 & 0.540
            & 0.426 & 0.392
            & 0.610 & 0.501 & 0.235
            & \underline{0.667} & 0.717 & \underline{0.600} & 0.775
            & \textbf{260.42} & \textbf{1.24$\times$} \\
        \bottomrule
    \end{tabular}%
    }
    \vspace{-0.1in}
\end{table}
\FloatBarrier

\section{Conclusion}

In this paper, we introduce \benchname{}, a benchmark for evaluating world memory in video world models through controlled departure-and-revisit rollouts. Across 260 cases, it separately assesses the retention of previously observed content, event evolution while out of view, general video quality, and adherence to prescribed camera trajectories. Evaluation of nine representative models shows that no model consistently leads across all dimensions and that strong static retention does not necessarily imply reliable dynamic state evolution. Multi-revisit evaluation reveals cumulative memory degradation beyond the first return, while cross-axis revisits expose additional sensitivity to changes in exploration direction. Human studies further support the alignment between automatic assessments and human judgments. With the model and generation settings held fixed, a controlled context-access study reveals how different temporal context policies affect memory performance and inference efficiency. Together, these findings establish \benchname{} as both an evaluation benchmark and a diagnostic tool for improving world memory. We hope that this benchmark will help advance video world models toward reliable simulators that support long-term interaction, learning, and planning in persistent virtual environments.

\bibliography{iclr2027_conference}
\bibliographystyle{iclr2027_conference}

\newpage
\appendix

\section{Overview of the Appendix}
In this section, we briefly introduce the contents of the appendix. We begin with a statement on LLM usage during manuscript preparation. 
In Sec.~\ref{app:inference-settings}, we detail the inference settings and output video specifications of the evaluated models. 
In Sec.~\ref{app:dynamic-memory-prompts}, we describe the Dynamic Memory assessment protocol and provide the corresponding prompt template. 
In Sec.~\ref{app:qualitative-results}, we present qualitative comparisons across Human, Object, and Environment cases.
In Sec.~\ref{app:human-eval-interface}, we describe the human evaluation protocols and annotation interfaces, and report the agreement between automatic assessments and human judgments. 
In Sec.~\ref{app:trajectory-results}, we provide trajectory-wise results and cross-axis comparisons.
In Sec.~\ref{app:context-policy-analysis}, we analyze how context-access policies affect static memory retention and dynamic state progression. 
Finally, in Sec.~\ref{app:limitation}, we discuss the limitations of \benchname{} and directions for future work.

\section{LLM Usage Statement}
\label{app:llm_usage}
During the development of this work, LLMs were used primarily to identify and correct grammatical errors and occasionally as rewriting tools to improve the clarity and readability of the text.

\section{Inference Settings}
\label{app:inference-settings}

All formal inference runs on machines equipped with eight NVIDIA H20-3e GPUs. Every model uses frozen random seeds and inference configurations, so that no setting varies across cases or trajectories.

\noindent \textbf{HY-WorldPlay}.
HY-WorldPlay consumes per-frame OpenCV camera-to-world poses; we extract one latent camera anchor every four pixel frames and generate its native pose condition with fixed camera intrinsics. Formal inference uses seed 1, 4-step distilled AR sampling, and eight-card model parallelism, with prompt rewriting and super-resolution disabled.

\noindent \textbf{Matrix-Game 3}.
The native interface of Matrix-Game 3 is jointly controlled by keyboard translation and mouse rotation. We compile the normalized camera trajectory into per-frame keyboard/mouse conditions. The model runs 12 iterations with seed 42 and 3-step AR sampling: the first segment generates 57 frames and each subsequent iteration adds 40 frames, producing 497 frames in total; inference uses seven GPUs for diffusion and one GPU for asynchronous MG-LightVAE decoding.

\noindent \textbf{Matrix-Game 3.5}.
Matrix-Game 3.5 constructs its native PRoPE condition from per-case camera intrinsics and camera-to-world poses. Formal inference adopts the released six-block distilled configuration, where each block corresponds to 84 camera poses, with a three-step schedule of $[1000,667,333]$ and seed 3407. The six blocks share the same content prompt and complete a continuous 505-frame rollout through the generated history and dynamic context.

\noindent \textbf{SANA-WM}.
SANA-WM builds chunk-level Pl\"ucker conditions from per-case camera intrinsics and poses, and generates videos through a streaming pipeline consisting of stage-1, a refiner, and a causal VAE. Stage-1 uses 4-step self-forcing sampling, and both stage-1 and the refiner treat three latent frames as one temporal block; formal inference fixes seed 42 and outputs 481 frames.

\noindent \textbf{YUME}.
YUME does not directly consume numerical camera poses. We convert the normalized trajectory into segmented camera captions composed of Left, Right, Up, Down, Hold, and translation directions; each caption stage corresponds to 29 frames, and the MemoryBench content prompt is appended verbatim to every stage. Formal inference disables additional prompt refinement and uses seed 43 with 4-step Euler sampling; LRLR and LRUD produce 493 frames, while LRLRLR, which contains three revisits, produces 551 frames.

\noindent \textbf{LingBot World}.
LingBot World interpolates per-frame camera-to-world poses to the latent rate and converts them into framewise relative poses and Pl\"ucker conditions; camera intrinsics are fixed on the model's input canvas. We set the chunk size to four latent frames so that the 493-frame target trajectory is generated in full, and run eight-card causal AR inference with seed 1 and FullKV attention.

\noindent \textbf{LingBot World v2}.
LingBot World v2 uses the same pose, intrinsics, and four-latent-frame chunk protocol as LingBot World. The nine-model main results use seed 1 and FullKV attention; FoV and Sliding Windows in Sec.~\ref{sec:lingbot-v2-context-management} are used only for the context-mechanism comparison with all other conditions frozen, and do not change the inference configuration behind the formal main table.

\noindent \textbf{Evoke}.
Evoke uses per-case camera intrinsics and a 501-frame absolute camera-to-world trajectory. The camera trajectory undergoes no additional smoothing or segmented motion cropping: absolute poses drive geometric warping, and latent-rate relative Pl\"ucker poses serve as the camera condition. Formal inference fixes seed 44, adopts 3-step CFG-free windowed AR generation, and keeps 21 static pose samples at the end to cover the final return view.

\noindent \textbf{DreamX World}.
DreamX World performs camera forcing with chunk-relative PRoPE conditions. Formal inference fixes seed 42, decodes 126 latent frames into 501 pixel frames, and retains the full generation history with FullKV attention.

\begin{table}
\vspace{-0.15in}
    \centering
    \caption{Video specifications and sampling configurations of the formal inference artifacts. Resolution and FPS are the models' native outputs; Frames / Duration reports the actual frame count of each output video together with its duration; Seed and Sampling give the sampling configuration of formal inference; Inference Setting only distinguishes explicit FullKV from the models' built-in memory modes.}
    \label{tab:inference-artifacts}
    \scriptsize
    \setlength{\tabcolsep}{6pt}
    \renewcommand{\arraystretch}{1.08}
    \begin{tabular}{@{}lcccccc@{}}
        \toprule
        Model & Resolution & FPS & Frames / Duration & Seed & Sampling & Inference Setting \\
        \midrule
        HY-WorldPlay     & $848{\times}480$  & 24 & 493 / 20.5 s             & 1    & 4-step AR            & Built-in \\
        Matrix-Game 3    & $1280{\times}704$ & 17 & 497 / 29.2 s             & 42   & 3-step AR $\times$12 & Built-in \\
        Matrix-Game 3.5  & $1280{\times}704$ & 16 & 505 / 31.6 s             & 3407 & 3-step, six-block    & Built-in \\
        SANA-WM          & $1280{\times}704$ & 16 & 481 / 30.1 s             & 42   & 4-step self-forcing  & FullKV \\
        YUME             & $1280{\times}704$ & 16 & 493--551 / 30.8--34.4 s & 43   & 4-step Euler         & Built-in \\
        LingBot World    & $832{\times}464$  & 16 & 493 / 30.8 s             & 1    & causal AR            & FullKV \\
        LingBot World v2 & $832{\times}464$  & 16 & 493 / 30.8 s             & 1    & causal AR            & FullKV \\
        Evoke            & $640{\times}384$  & 24 & 501 / 20.9 s             & 44   & 3-step windowed AR   & Built-in \\
        DreamX World     & $1280{\times}704$ & 16 & 501 / 31.3 s             & 42   & distilled AR         & FullKV \\
        \bottomrule
    \end{tabular}
\end{table}

\section{Dynamic Memory Evaluation Details}
\label{app:dynamic-memory-prompts}

\subsection{Assessment Protocol}
Each rollout is sampled at a fixed stride of 8 frames, always including the first and last frames. The sampled frames are provided to Qwen3.8-27B in chronological order, together with the checklist and the planned camera-phase ranges (initial hold, departures, revisits, and final stationary hold). These ranges allow the assessment model to locate the evidence window for each question without access to the generation prompt. The model first determines whether the central event visibly starts during the initial hold and then answers all eight questions. If the model determines that the event has not visibly started, all eight question scores are set to zero. Otherwise, each question receives one point if the model's answer matches the frozen reference answer. We set \texttt{max\_tokens} to 4096 for the assessment model.

\subsection{Event Initiation and Conditional Dynamic Scores}
\label{app:dynamic-event-decomposition}
We separate the event-start prerequisite from subsequent checklist performance
using the same 60 videos per model.
The event-start rate $p_{\mathrm{start}}$ is the fraction of videos judged
to show concrete event progress during the initial hold.
The conditional score $S_{\mathrm{cond}}$ averages the eight-question
checklist score only over those videos. Because videos with a negative gate
receive zero, the reported Overall satisfies
$S_{\mathrm{gated}}=p_{\mathrm{start}}S_{\mathrm{cond}}$.

\begin{table}[t]
\centering
\caption{Dynamic Memory decomposition from the frozen 60-case evaluation.
Started is the number of judge-positive event-start decisions. Conditional averages only the Started subset.}
\label{tab:dynamic-event-decomposition}
\small
\setlength{\tabcolsep}{5pt}
\begin{tabular}{@{}lrrrr@{}}
\toprule
Model & Started & Start rate (\%) & Gated Overall & Conditional\\
\midrule
HY-WorldPlay & 19/60 & 31.7 & 0.294 & 0.928 \\
Matrix-Game 3 & 3/60 & 5.0 & 0.046 & 0.917 \\
Matrix-Game 3.5 & 23/60 & 38.3 & 0.240 & 0.625 \\
SANA-WM & 35/60 & 58.3 & 0.415 & 0.711 \\
YUME & 54/60 & 90.0 & 0.775 & 0.861 \\
LingBot World & 56/60 & 93.3 & 0.808 & 0.866 \\
LingBot World v2 & 53/60 & 88.3 & 0.696 & 0.788 \\
Evoke & 49/60 & 81.7 & 0.706 & 0.865 \\
DreamX World & 39/60 & 65.0 & 0.563 & 0.865 \\
\bottomrule
\end{tabular}
\end{table}

HY-WorldPlay has a conditional score of 0.928 on 19 started events,
but its 31.7\% start rate yields a gated Overall of 0.294.
LingBot World starts 56 of 60 events and obtains a conditional score of
0.866, yielding 0.808 overall.
Thus, differences in the headline Dynamic Memory scores reflect both
event initiation and subsequent checklist performance.
Matrix-Game 3's conditional score of 0.917 is based on only three videos.
We retain the gated score as the end-to-end benchmark result and use the
decomposition to distinguish these sources of failure.
The conditional scores remain judge-based evaluations of different
successful-start subsets, not isolated measures of memory.

\subsection{Assessment Prompt}
Below, we present the prompt template used for Dynamic Memory assessment.

\begin{promptbox}{Dynamic Memory Assessment Prompt}
You are judging one generated Dynamic Memory LR-hold video using eight case-specific questions. The attached images are sampled in chronological order. The target generation prompt and expected answers are deliberately hidden.

Use only visible evidence; do not complete a plausible story from the question wording. Question wording describes the intended evaluation target, but it is not visual evidence. For a change claim, cite visible before-and-after images that show a concrete difference; otherwise answer no. If the required subject or object disappears, is replaced, or changes into a different scene, do not treat its absence as successful completion. In the reason, describe only concrete evidence visible in the cited images and do not invent an unseen state or event.

The evaluator has assigned the images to the model's planned camera phases. Each range below is an inclusive pair of 1-based image numbers.

\textbf{Planned phase image ranges:} <planned phase ranges JSON: initial\_hold, departures, revisits, final\_stationary>

\textbf{Useful comparison ranges:} <compact comparison ranges JSON>

\textbf{Phase guidance:}\\
- returned\_view means the returned\_hold range, compared with the departure\_baseline range when the question asks about change.\\
- returned\_hold\_sequence means the progression inside the returned\_hold range.\\
- returned\_and\_later\_visible\_sequence means the returned\_hold range through its final images.\\
- final\_stationary\_view means the final\_stationary range.\\
- ordered\_visible\_sequence may use the full chronological image sequence.\\
- For a comparison or temporal claim, use visible before/after evidence when available.\\
- If an event never occurs, say that directly. Never invent a reset, reversal, worsening, disappearance, or change that is not visible.\\
- Interpret extent words relative to the stated baseline, not by absolute impression.\\
- Camera motion, framing, lighting, an action pose, or a replaced subject alone is not evidence of state progress.\\
- If the exact state is not visibly established, answer no.\\
- Cite only decisive 1-based image numbers.

\textbf{Event-start prerequisite:}\\
- First decide whether the central event described by the questions visibly starts during initial\_hold, before the camera leaves the original view.\\
- event\_started is yes only when at least two initial\_hold images show concrete temporal progress by the principal subject or object.\\
- Mere pose, contact, camera motion, background motion, or a plausible intention is not an event start.\\
- If no relevant action or state change is visibly underway by the end of initial\_hold, event\_started is no.\\
- Even when event\_started is no, answer all eight questions from visible evidence; the evaluator applies the zero-score gate.

\textbf{Questions (expected answers are intentionally hidden):} <checklist JSON: id, category, evidence\_window, question for each of the eight frozen questions>

\textbf{Output format.} Return one JSON object with exactly two keys:\\
- event\_started: \{answer: "yes" or "no"; evidence\_image\_numbers: at least two decisive 1-based image numbers from initial\_hold; reason: one concise sentence describing the visible change, or the visible lack of change\}.\\
- answers: exactly eight objects in the same ID order, each \{id: the question ID; answer: "yes" or "no"; evidence\_image\_numbers: a short list of decisive 1-based image numbers; reason: one concise sentence describing only the cited visible evidence\}.\\
Return JSON only.
\end{promptbox}

% \clearpage
\section{Qualitative Results}
\label{app:qualitative-results}
In this section, we present Human and Object qualitative comparisons of the video world models evaluated in
Sec.~\ref{sec:Evaluation}, complementing the Environment case in Fig.~\ref{fig:vis_3} in the main text. Target regions in the Human and Object cases are
highlighted with red boxes, and the corresponding metric scores are shown for
each revisit. Across all three cases, HY-WorldPlay provides the most consistent
multi-revisit recovery: it preserves the subject's identity and appearance, the
wardrobe's appearance and geometry, and the room's visual content and spatial
structure through $R_3$. This observation matches Tab.~\ref{tab:main-results},
where HY-WorldPlay leads all seven Static Memory metrics. Evoke also remains
comparatively stable in the Human and Object examples, consistent with its
second-best results on the four entity-level metrics, while other models show
identity drift, target disappearance, or scene-layout changes at later
revisits.

\begin{figure}
    \vspace{-0.15in}
    \centering
    \includegraphics[width=\linewidth]{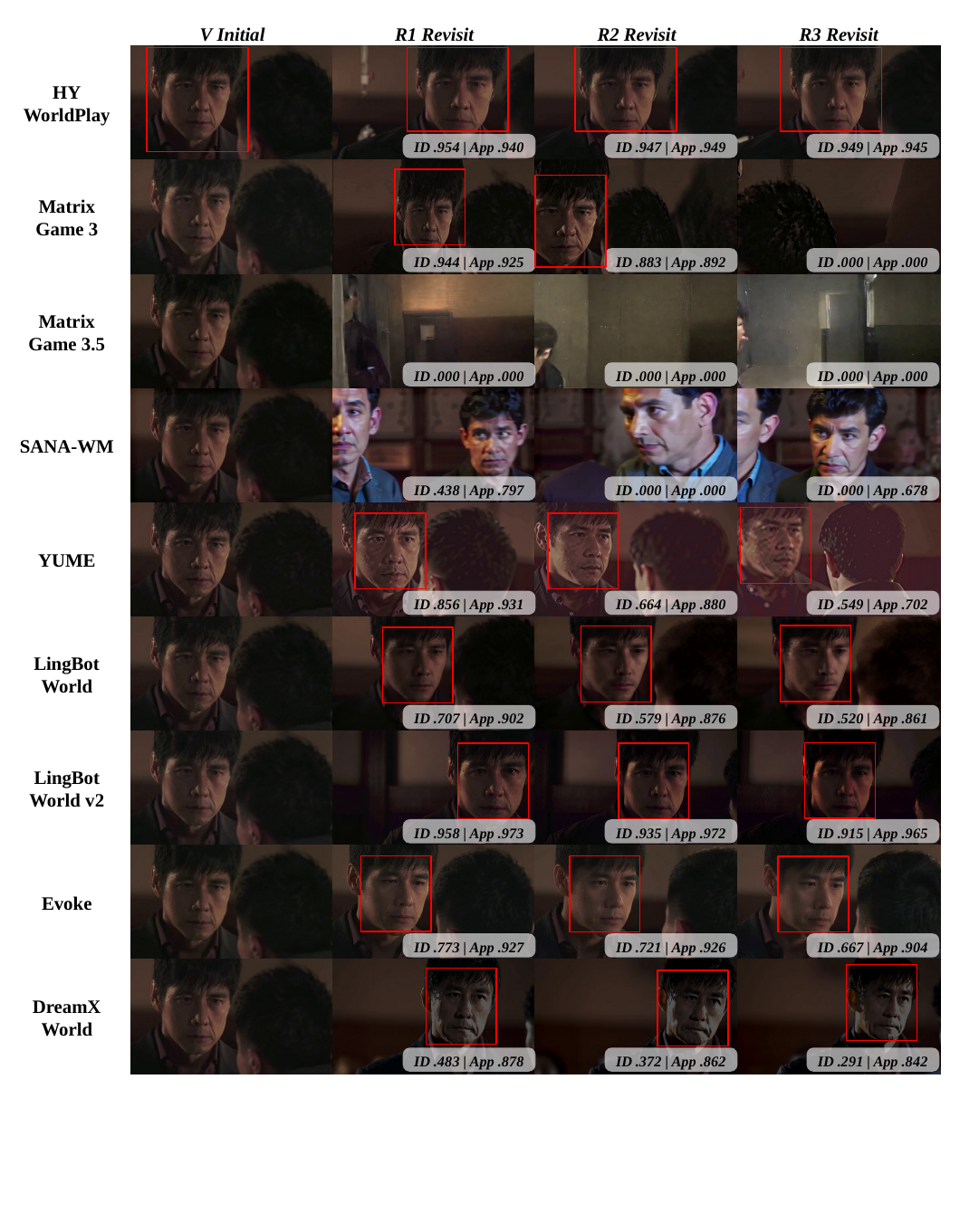}
    \vspace{-0.25in}
    \caption{
        Qualitative comparison of video world models on a human case.
    }
    \label{fig:vis_1}
    \vspace{-0.15in}
\end{figure}

\begin{figure}
    \vspace{-0.15in}
    \centering
    \includegraphics[width=\linewidth]{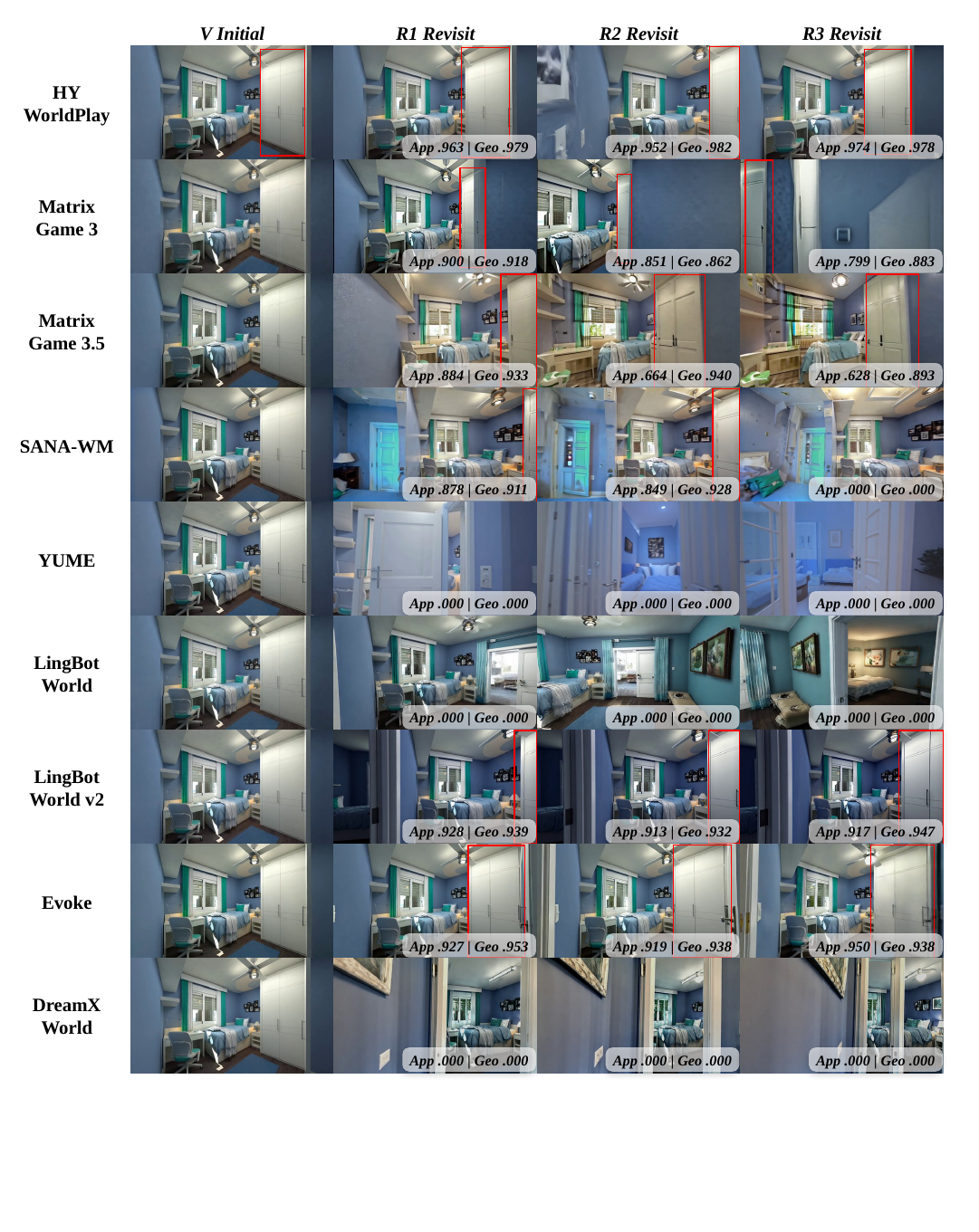}
    \vspace{-0.25in}
    \caption{
        Qualitative comparison of video world models on an object case.
    }
    \label{fig:vis_2}
    \vspace{-0.15in}
\end{figure}

\FloatBarrier

\section{Human Evaluation}
\label{app:human-eval-interface}

\noindent \textbf{Static Memory protocol.}
We conduct a blinded A/B preference study on whether the six Static Memory
metrics reproduce human judgments of long-term memory retention. In total, 504
blinded pairs are each judged by three annotators as A-wins, Tie, or B-wins,
with model identities and order hidden, yielding 1{,}512 judgements (252 per
metric). Pairwise agreement between the automatic metric and the human
majority-vote direction serves as the primary measure; we further compare
model-level win rates (Win/Tie/Loss as $1/0.5/0$) via Spearman $\rho$.

Pairwise agreement compares the automatic preference with a decisive
human majority. Majority ties and three-way A/Tie/B splits are excluded
from this denominator and reported separately.
For example, Human Identity has 65 agreements among 76 decisive pairs,
giving 85.5\%; one majority tie and seven three-way splits account for
the remaining eight pairs.
Model-level Spearman $\rho$ instead uses all raw votes to compute human
win rates (Win/Tie/Loss as $1/0.5/0$), compared with automatic win rates
on the same 84 tasks. 

All six metrics reach 78.5\%--92.5\% pairwise agreement with human majority votes (Tab.~\ref{tab:static-human-alignment}). At the model-ranking level, Environment Visible, Human Appearance, and Human Identity attain $\rho=0.988$, $0.952$, and $0.929$, confirming that the automatic metrics capture human-perceivable differences in long-term memory retention.

\begin{table}[h]
\centering
% \vspace{-0.15in}
\small
\caption{Agreement between automatic Static Memory metrics and blinded human A/B preferences. Pairwise Agreement measures the per-pair directional consistency between an automatic metric and the human majority vote; model-level Spearman $\rho$ is computed over the human and automatic win rates of the eight models.}
\label{tab:static-human-alignment}
\setlength{\tabcolsep}{1.8mm}
\begin{tabular}{@{}lrrrr@{}}
    \toprule
    Metric & Agree/Decisive & Tie/Split & Agreement (\%) & Spearman $\rho$ \\
    \midrule
    Human Identity      & 65/76 & 1/7 & 85.5 & 0.929 \\
    Human Appearance    & 71/80 & 2/2 & 88.8 & 0.952 \\
    Object Geometry     & 62/79 & 0/5 & 78.5 & 0.606 \\
    Object Appearance   & 67/80 & 2/2 & 83.8 & 0.699 \\
    Environment Visible & 74/80 & 1/3 & 92.5 & 0.988 \\
    Environment Depth   & 63/77 & 4/3 & 81.8 & 0.301 \\
    \bottomrule
\end{tabular}
\end{table}

\begin{figure}[h]
    \centering
    \includegraphics[width=0.92\textwidth]{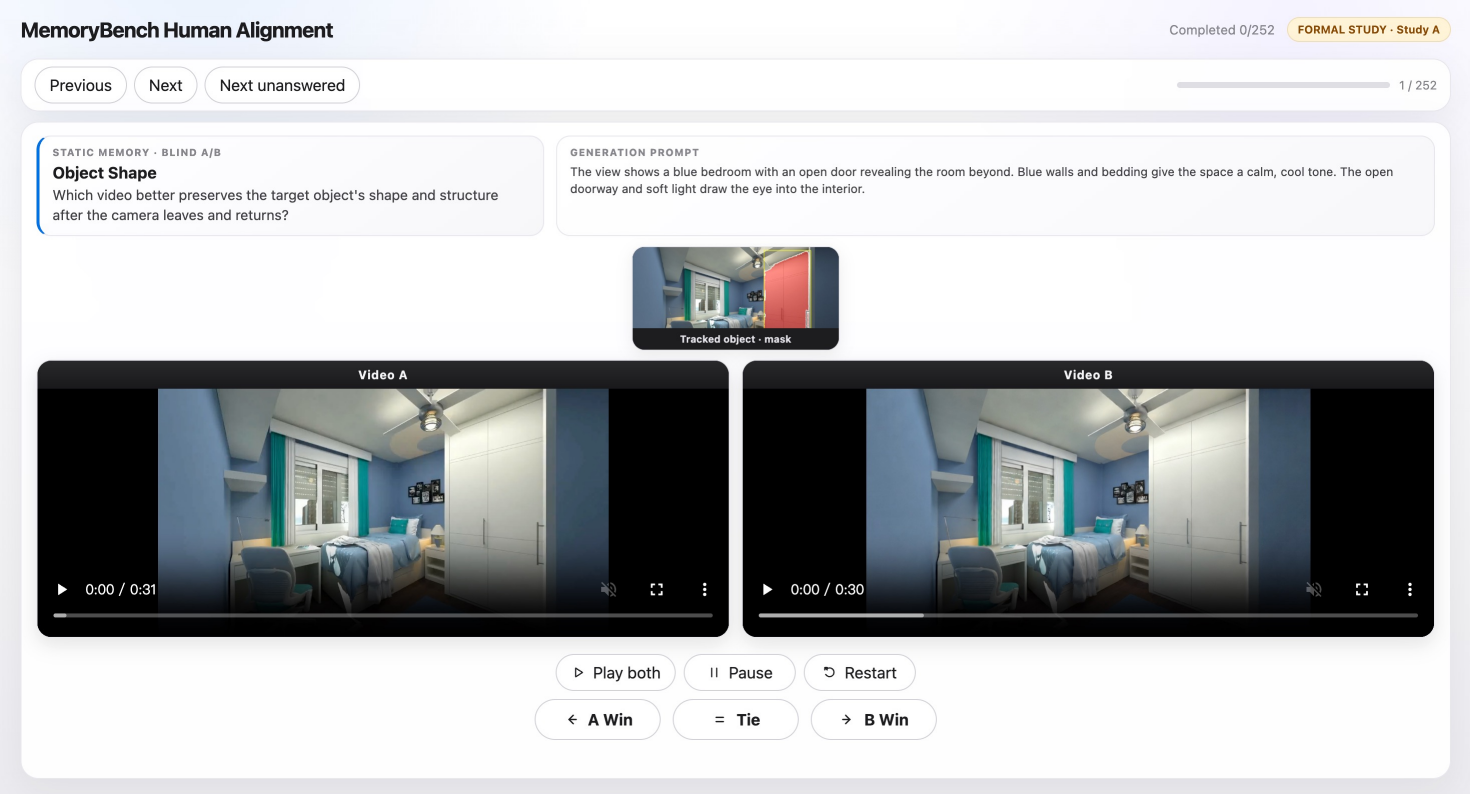}
    \caption{Blinded A/B preference annotation interface for Static Memory.}
    \label{fig:human-eval-static}
\end{figure}

\noindent \textbf{Dynamic Memory protocol.}
We validate the Dynamic Memory VLM judge on 50 evaluation instances with 400 checklist items. Three annotators independently answer the eight binary questions of each instance under the same event gate as the judge: if the target event has not started, all items of that instance are labeled 0. Each item thus receives three judgements (1{,}200 in total), whose majority vote forms the human reference label; human and VLM scores are the respective fractions of satisfied checklist items per instance.

The judge attains an Overall Spearman $\rho=0.732$ and PLCC of $0.808$ against human evaluation (Tab.~\ref{tab:dynamic-human-alignment}). Across the four sub-dimensions, $\rho$ spans 0.401--0.684 and PLCC spans 0.491--0.683, with the highest agreement on Return Progress ($\rho=0.684$, PLCC $=0.683$). These results validate checklist-based Dynamic Memory evaluation as an effective and scalable automatic protocol.

\begin{table}[h]
    \centering
    \vspace{-0.15in}
    \small
    \caption{Human alignment of the Dynamic Memory checklist evaluation. The per-item majority vote of three annotators forms the human reference labels; Spearman $\rho$ and PLCC are computed over the checklist satisfaction scores of the 50 instances.}
    \label{tab:dynamic-human-alignment}
    \setlength{\tabcolsep}{5mm}
    \begin{tabular}{lcccc}
        \toprule
        Task & Checklist Items & Judgements & Spearman $\rho\uparrow$ & PLCC $\uparrow$ \\
        \midrule
        Return Progress       & 100  & 300 & 0.684 & 0.683 \\
        Persistence / Reset   & 100  & 300 & 0.582 & 0.612 \\
        Final State           & 100  & 300 & 0.624 & 0.641 \\
        Causal / Physics      & 100  & 300 & 0.401 & 0.491 \\
        \midrule
        Overall               & 400 & 1200 & 0.732 & 0.808 \\
        \bottomrule
    \end{tabular}
\end{table}

\begin{figure}[h]
    \centering
    \includegraphics[width=0.92\textwidth]{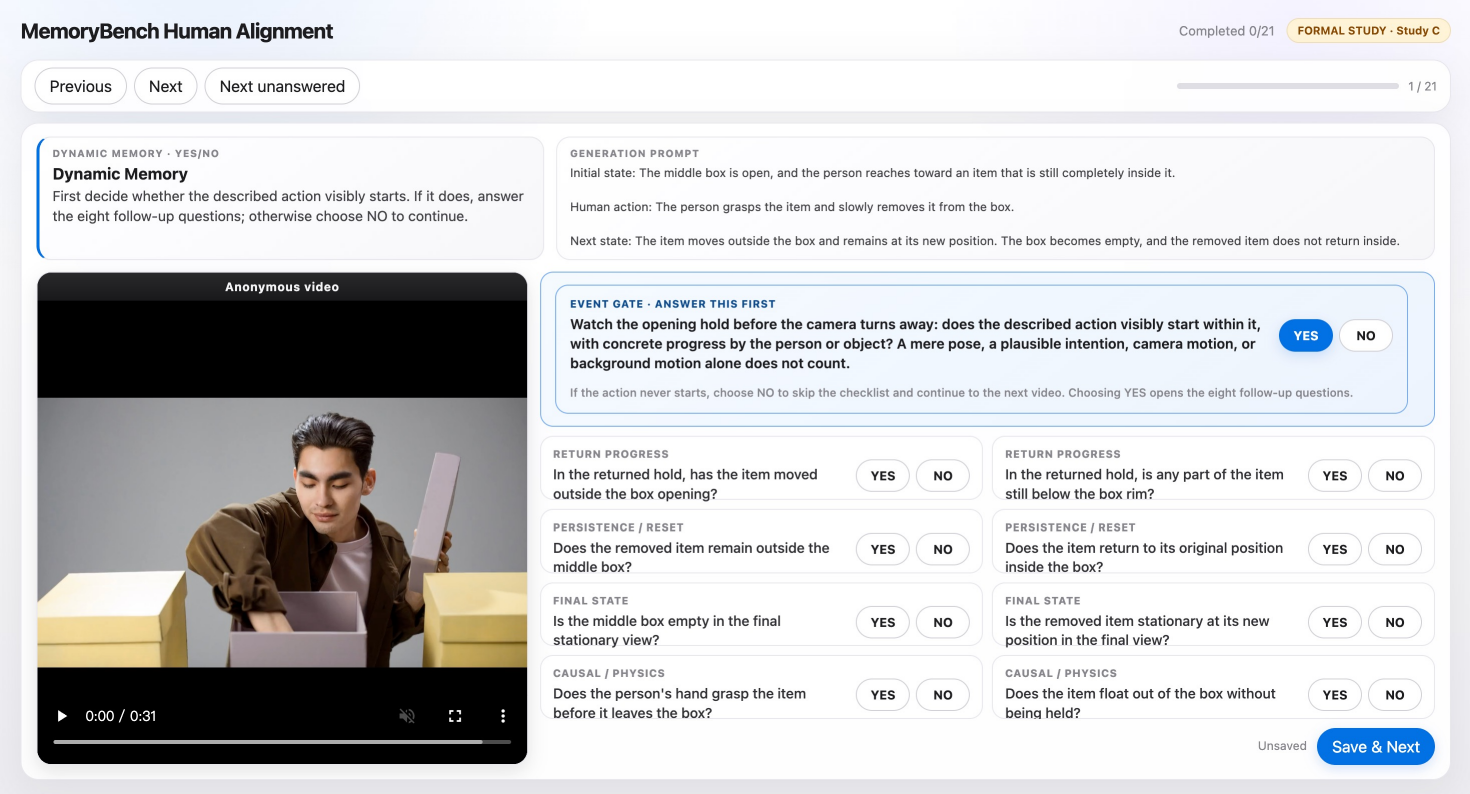}
    \caption{Event-gated checklist annotation interface for Dynamic Memory.}
    \label{fig:human-eval-dynamic}
\end{figure}

\section{Trajectory-Wise and Cross-Axis Results}
\label{app:trajectory-results}

% This section supplements the aggregate main table with per-model results for
% LRLR, LRLRLR, and LRUD, followed by the LRLR--LRUD comparison underlying the
% cross-axis analysis in Section~\ref{sec:cross-axis-revisit-analysis}.
% Environment Reconstruction is reported as a higher-is-better score throughout.

% \input{sections/generated_trajectory_result_tables}
% \input{sections/generated_cross_axis_summary_table}
% \input{sections/generated_cross_axis_revisit_table}
% \FloatBarrier

In this section, we supplement the aggregate main results with per-model results for LRLR, LRLRLR, and LRUD, followed by the LRLR--LRUD comparison underlying the cross-axis analysis in Sec.~\ref{sec:cross-axis-revisit-analysis}. Environment Reconstruction is reported as a higher-is-better score throughout.

\noindent \textbf{Trajectory-wise results.}
Tab.~\ref{tab:trajectory-per-model} shows the results of all nine models
on the three rotation trajectories.
HY-WorldPlay achieves the highest scores on all seven Static Memory
metrics under LRLR and LRLRLR, and on six under LRUD, where Evoke leads
in Human Appearance.
Meanwhile, YUME achieves the highest Imaging scores across all three
trajectories, indicating that strong visual quality does not necessarily
coincide with the strongest memory retention.

% BEGIN inlined sections/generated_trajectory_result_tables.tex
\begin{table}[h]
\vspace{-0.15in}
    \centering
    \caption{Per-model results on the three formal Rotation trajectories. Best and second-best results within each trajectory are bold and underlined.}
    \label{tab:trajectory-per-model}
    \scriptsize
    \setlength{\tabcolsep}{2.0pt}
    \renewcommand{\arraystretch}{0.95}
    \resizebox{\textwidth}{!}{%
    \begin{tabular}{@{}l*{9}{c}@{}}
        \toprule
        Metric & HY-WorldPlay & Matrix-Game 3 & Matrix-Game 3.5 & SANA-WM & YUME & LingBot World & LingBot World v2 & Evoke & DreamX World \\
        \midrule
        \multicolumn{10}{@{}l}{\emph{(a) LRLR}} \\
        Imaging $\uparrow$ & 0.668 & 0.686 & 0.613 & 0.554 & \textbf{0.712} & 0.677 & \underline{0.698} & 0.673 & 0.683 \\
        Aesthetic $\uparrow$ & 0.538 & 0.522 & 0.446 & 0.451 & 0.539 & \textbf{0.556} & \underline{0.553} & 0.541 & 0.530 \\
        Human ID $\uparrow$ & \textbf{0.863} & \underline{0.577} & 0.015 & 0.150 & 0.295 & 0.389 & 0.559 & 0.546 & 0.267 \\
        Human App. $\uparrow$ & \textbf{0.889} & 0.597 & 0.041 & 0.282 & 0.529 & 0.643 & 0.702 & \underline{0.783} & 0.522 \\
        Object App. $\uparrow$ & \textbf{0.925} & 0.647 & 0.510 & 0.311 & 0.498 & 0.442 & 0.626 & \underline{0.744} & 0.486 \\
        Object Geometry $\uparrow$ & \textbf{0.847} & 0.547 & 0.501 & 0.303 & 0.411 & 0.422 & 0.559 & \underline{0.689} & 0.421 \\
        Env. Visible $\uparrow$ & \textbf{0.953} & \underline{0.818} & 0.677 & 0.529 & 0.691 & 0.787 & 0.700 & 0.746 & 0.563 \\
        Env. Depth $\uparrow$ & \textbf{0.920} & 0.640 & 0.666 & 0.663 & 0.689 & \underline{0.743} & 0.515 & 0.697 & 0.378 \\
        Env. Recon. $\uparrow$ & \textbf{0.657} & \underline{0.460} & 0.411 & 0.342 & 0.378 & 0.440 & 0.281 & 0.404 & 0.073 \\
        Rot. RPE $\downarrow$ & \textbf{0.366} & 0.373 & 0.656 & 0.429 & 0.506 & 0.586 & \underline{0.371} & 0.556 & 0.961 \\
        \midrule
        \multicolumn{10}{@{}l}{\emph{(b) LRLRLR}} \\
        Imaging $\uparrow$ & 0.687 & 0.688 & 0.606 & 0.511 & \textbf{0.711} & 0.672 & 0.690 & 0.670 & \underline{0.691} \\
        Aesthetic $\uparrow$ & \textbf{0.559} & 0.522 & 0.445 & 0.432 & 0.539 & \underline{0.554} & 0.550 & 0.539 & 0.549 \\
        Human ID $\uparrow$ & \textbf{0.850} & \underline{0.602} & 0.026 & 0.140 & 0.339 & 0.448 & 0.547 & 0.548 & 0.333 \\
        Human App. $\uparrow$ & \textbf{0.875} & 0.624 & 0.057 & 0.284 & 0.650 & 0.720 & 0.673 & \underline{0.742} & 0.619 \\
        Object App. $\uparrow$ & \textbf{0.900} & 0.679 & 0.540 & 0.485 & 0.686 & 0.512 & 0.576 & 0.676 & \underline{0.688} \\
        Object Geometry $\uparrow$ & \textbf{0.772} & 0.595 & 0.540 & 0.464 & 0.590 & 0.471 & 0.542 & \underline{0.634} & 0.592 \\
        Env. Visible $\uparrow$ & \textbf{0.956} & \underline{0.868} & 0.699 & 0.661 & 0.767 & 0.797 & 0.693 & 0.797 & 0.708 \\
        Env. Depth $\uparrow$ & \textbf{0.924} & 0.689 & 0.724 & 0.731 & 0.695 & \underline{0.783} & 0.524 & 0.759 & 0.515 \\
        Env. Recon. $\uparrow$ & \textbf{0.712} & \underline{0.518} & 0.478 & 0.473 & 0.358 & 0.409 & 0.274 & 0.430 & 0.174 \\
        Rot. RPE $\downarrow$ & 0.453 & \textbf{0.282} & 0.511 & 0.393 & 0.532 & 0.538 & \underline{0.348} & 0.541 & 0.825 \\
        \midrule
        \multicolumn{10}{@{}l}{\emph{(c) LRUD}} \\
        Imaging $\uparrow$ & 0.621 & 0.667 & 0.592 & 0.559 & \textbf{0.715} & 0.664 & 0.662 & 0.667 & \underline{0.686} \\
        Aesthetic $\uparrow$ & 0.486 & 0.487 & 0.395 & 0.419 & 0.521 & \underline{0.527} & 0.500 & 0.518 & \textbf{0.531} \\
        Human ID $\uparrow$ & \textbf{0.771} & 0.272 & 0.015 & 0.110 & 0.259 & 0.343 & 0.494 & \underline{0.557} & 0.259 \\
        Human App. $\uparrow$ & \underline{0.727} & 0.285 & 0.030 & 0.188 & 0.473 & 0.619 & 0.560 & \textbf{0.763} & 0.528 \\
        Object App. $\uparrow$ & \textbf{0.814} & 0.510 & 0.363 & 0.186 & 0.409 & 0.390 & 0.491 & \underline{0.710} & 0.534 \\
        Object Geometry $\uparrow$ & \textbf{0.664} & 0.413 & 0.329 & 0.175 & 0.345 & 0.362 & 0.417 & \underline{0.628} & 0.459 \\
        Env. Visible $\uparrow$ & \textbf{0.933} & 0.661 & 0.483 & 0.421 & 0.589 & 0.601 & 0.690 & \underline{0.743} & 0.559 \\
        Env. Depth $\uparrow$ & \textbf{0.871} & 0.504 & 0.545 & 0.610 & 0.644 & 0.514 & 0.513 & \underline{0.734} & 0.363 \\
        Env. Recon. $\uparrow$ & \textbf{0.488} & 0.292 & 0.282 & 0.287 & 0.303 & 0.242 & 0.205 & \underline{0.367} & 0.066 \\
        Rot. RPE $\downarrow$ & \textbf{0.433} & 1.008 & 1.343 & 0.569 & \underline{0.549} & 0.786 & 0.598 & 0.622 & 0.997 \\
        \bottomrule
    \end{tabular}%
    }
\end{table}
% END inlined sections/generated_trajectory_result_tables.tex

\noindent \textbf{Average cross-axis degradation.}
Tab.~\ref{tab:cross-axis-summary} summarizes the nine-model mean scores at the two revisits, with $d=R_1-R_2$ measuring revisit-to-revisit degradation and $\Delta_{\mathrm{axis}}=d_{\mathrm{LRUD}}-d_{\mathrm{LRLR}}$ measuring the excess degradation associated with switching rotation axes. Both trajectories show lower mean scores at the second revisit across all seven metrics, with larger drops under LRUD. The largest excess degradation occurs in Environment Visible ($+0.152$), followed by Object Appearance ($+0.127$) and Object Geometry ($+0.120$).

% \input{sections/generated_cross_axis_summary_table}
% BEGIN inlined sections/generated_cross_axis_summary_table.tex
\begin{table}[h]
\vspace{-0.15in}
    \centering
    \caption{Nine-model mean Static Memory scores at the two revisits of LRLR and LRUD. $d=R_1-R_2$ and $\Delta_{\mathrm{axis}}=d_{\mathrm{LRUD}}-d_{\mathrm{LRLR}}$; positive $\Delta_{\mathrm{axis}}$ indicates greater degradation after switching from yaw to pitch.}
    \label{tab:cross-axis-summary}
    \small
    \setlength{\tabcolsep}{3.3pt}
    \begin{tabular}{@{}lccccccc@{}}
        \toprule
        \multirow{2}{*}{Metric} & \multicolumn{3}{c}{LRLR} & \multicolumn{3}{c}{LRUD} & \multirow{2}{*}{$\Delta_{\mathrm{axis}}$} \\
        \cmidrule(lr){2-4}\cmidrule(lr){5-7}
        & $R_1$ & $R_2$ & $d$ & $R_1$ & $R_2$ & $d$ & \\
        \midrule
        Human ID $\uparrow$ & 0.490 & 0.324 & 0.166 & 0.435 & 0.249 & 0.186 & +0.020 \\
        Human App. $\uparrow$ & 0.623 & 0.486 & 0.137 & 0.556 & 0.371 & 0.185 & +0.049 \\
        Object App. $\uparrow$ & 0.657 & 0.496 & 0.162 & 0.634 & 0.345 & 0.288 & +0.127 \\
        Object Geometry $\uparrow$ & 0.588 & 0.457 & 0.132 & 0.547 & 0.296 & 0.251 & +0.120 \\
        Env. Visible $\uparrow$ & 0.781 & 0.655 & 0.126 & 0.770 & 0.492 & 0.278 & \textbf{+0.152} \\
        Env. Depth $\uparrow$ & 0.703 & 0.611 & 0.091 & 0.688 & 0.489 & 0.199 & +0.107 \\
        Env. Recon. $\uparrow$ & 0.454 & 0.312 & 0.141 & 0.386 & 0.177 & 0.209 & +0.068 \\
        \bottomrule
    \end{tabular}
\end{table}
% END inlined sections/generated_cross_axis_summary_table.tex

\noindent \textbf{Per-model cross-axis comparison.}
Tab.~\ref{tab:cross-axis-revisits} provides the complete per-model
revisit scores and cross-axis differences.
All nine models exhibit positive $\Delta_{\mathrm{axis}}$ on Environment
Visible, and eight of nine exhibit positive values on both Object
Appearance and Object Geometry.
The magnitude nevertheless varies substantially across models:
LingBot World shows excess degradation of $+0.368$ on Environment Visible
and $+0.465$ on Environment Depth, whereas Evoke shows $+0.002$ and
$-0.098$, respectively.
These results indicate that the average cross-axis trend masks
substantial differences in sensitivity across models and metrics.

\clearpage

% \input{sections/generated_cross_axis_revisit_table}
% \input{sections/generated_cross_axis_revisit_table}
% BEGIN inlined sections/generated_cross_axis_revisit_table.tex
\begin{table}[h]
\vspace{-0.15in}
    \centering
    \caption{Full per-model LRLR--LRUD cross-axis comparison. Panels (a) and (b) report the two revisit scores; panel (c) reports $\Delta_{\mathrm{axis}}$. All memory metrics are higher-is-better.}
    \label{tab:cross-axis-revisits}
    \scriptsize
    \setlength{\tabcolsep}{1.8pt}
    \renewcommand{\arraystretch}{0.96}
    \resizebox{\textwidth}{!}{%
    \begin{tabular}{@{}ll*{9}{c}c@{}}
        \toprule
        \multirow{2}{*}{\textbf{Metric}} & \multirow{2}{*}{\textbf{Slot}}
        & \textbf{HY} & \textbf{Matrix} & \textbf{Matrix}
        & \multirow{2}{*}{\textbf{SANA-WM}} & \multirow{2}{*}{\textbf{YUME}}
        & \textbf{LingBot} & \textbf{LingBot} & \multirow{2}{*}{\textbf{Evoke}}
        & \textbf{DreamX} & \multirow{2}{*}{\textbf{Mean}} \\
        & & \textbf{WorldPlay} & \textbf{Game-3} & \textbf{Game-3.5}
        & & & \textbf{World} & \textbf{World-v2} & & \textbf{World} & \\
        \midrule
        \multicolumn{12}{@{}l}{\emph{(a) LRLR: two yaw revisits}} \\
        \multirow{2}{*}{Human ID $\uparrow$} & $R1$ & 0.873 & 0.738 & 0.030 & 0.229 & 0.431 & 0.517 & 0.663 & 0.579 & 0.349 & 0.490 \\
         & $R2$ & 0.853 & 0.416 & 0.000 & 0.072 & 0.159 & 0.261 & 0.456 & 0.514 & 0.186 & 0.324 \\
        \multirow{2}{*}{Human App. $\uparrow$} & $R1$ & 0.894 & 0.719 & 0.076 & 0.395 & 0.620 & 0.734 & 0.785 & 0.799 & 0.582 & 0.623 \\
         & $R2$ & 0.884 & 0.476 & 0.006 & 0.168 & 0.438 & 0.553 & 0.618 & 0.767 & 0.463 & 0.486 \\
        \multirow{2}{*}{Object App. $\uparrow$} & $R1$ & 0.928 & 0.700 & 0.694 & 0.324 & 0.596 & 0.605 & 0.661 & 0.745 & 0.663 & 0.657 \\
         & $R2$ & 0.922 & 0.593 & 0.327 & 0.299 & 0.399 & 0.279 & 0.592 & 0.743 & 0.308 & 0.496 \\
        \multirow{2}{*}{Object Geometry $\uparrow$} & $R1$ & 0.859 & 0.589 & 0.665 & 0.304 & 0.483 & 0.553 & 0.603 & 0.696 & 0.543 & 0.588 \\
         & $R2$ & 0.835 & 0.506 & 0.337 & 0.302 & 0.339 & 0.291 & 0.516 & 0.683 & 0.300 & 0.457 \\
        \multirow{2}{*}{Env. Visible $\uparrow$} & $R1$ & 0.959 & 0.830 & 0.795 & 0.654 & 0.763 & 0.825 & 0.787 & 0.777 & 0.642 & 0.781 \\
         & $R2$ & 0.946 & 0.806 & 0.560 & 0.404 & 0.619 & 0.748 & 0.613 & 0.714 & 0.484 & 0.655 \\
        \multirow{2}{*}{Env. Depth $\uparrow$} & $R1$ & 0.937 & 0.674 & 0.753 & 0.731 & 0.726 & 0.781 & 0.594 & 0.730 & 0.397 & 0.703 \\
         & $R2$ & 0.903 & 0.606 & 0.580 & 0.596 & 0.652 & 0.704 & 0.435 & 0.664 & 0.360 & 0.611 \\
        \multirow{2}{*}{Env. Recon. $\uparrow$} & $R1$ & 0.723 & 0.478 & 0.490 & 0.484 & 0.450 & 0.482 & 0.419 & 0.443 & 0.111 & 0.454 \\
         & $R2$ & 0.591 & 0.442 & 0.331 & 0.200 & 0.306 & 0.397 & 0.142 & 0.364 & 0.035 & 0.312 \\
        \midrule
        \multicolumn{12}{@{}l}{\emph{(b) LRUD: yaw followed by pitch}} \\
        \multirow{2}{*}{Human ID $\uparrow$} & $R1$ & 0.840 & 0.364 & 0.031 & 0.205 & 0.427 & 0.504 & 0.633 & 0.580 & 0.337 & 0.435 \\
         & $R2$ & 0.702 & 0.180 & 0.000 & 0.016 & 0.091 & 0.181 & 0.356 & 0.535 & 0.182 & 0.249 \\
        \multirow{2}{*}{Human App. $\uparrow$} & $R1$ & 0.828 & 0.396 & 0.060 & 0.341 & 0.593 & 0.729 & 0.735 & 0.742 & 0.580 & 0.556 \\
         & $R2$ & 0.625 & 0.174 & 0.000 & 0.035 & 0.354 & 0.508 & 0.384 & 0.783 & 0.475 & 0.371 \\
        \multirow{2}{*}{Object App. $\uparrow$} & $R1$ & 0.876 & 0.667 & 0.633 & 0.310 & 0.598 & 0.593 & 0.634 & 0.740 & 0.651 & 0.634 \\
         & $R2$ & 0.751 & 0.353 & 0.093 & 0.062 & 0.219 & 0.186 & 0.347 & 0.681 & 0.417 & 0.345 \\
        \multirow{2}{*}{Object Geometry $\uparrow$} & $R1$ & 0.754 & 0.534 & 0.583 & 0.292 & 0.475 & 0.542 & 0.552 & 0.652 & 0.541 & 0.547 \\
         & $R2$ & 0.574 & 0.293 & 0.075 & 0.059 & 0.215 & 0.183 & 0.283 & 0.604 & 0.377 & 0.296 \\
        \multirow{2}{*}{Env. Visible $\uparrow$} & $R1$ & 0.951 & 0.829 & 0.767 & 0.598 & 0.764 & 0.823 & 0.780 & 0.775 & 0.645 & 0.770 \\
         & $R2$ & 0.914 & 0.493 & 0.200 & 0.244 & 0.414 & 0.378 & 0.600 & 0.710 & 0.473 & 0.492 \\
        \multirow{2}{*}{Env. Depth $\uparrow$} & $R1$ & 0.898 & 0.648 & 0.726 & 0.688 & 0.737 & 0.785 & 0.597 & 0.718 & 0.394 & 0.688 \\
         & $R2$ & 0.845 & 0.360 & 0.363 & 0.532 & 0.551 & 0.243 & 0.429 & 0.750 & 0.331 & 0.489 \\
        \multirow{2}{*}{Env. Recon. $\uparrow$} & $R1$ & 0.541 & 0.470 & 0.369 & 0.354 & 0.424 & 0.482 & 0.294 & 0.433 & 0.106 & 0.386 \\
         & $R2$ & 0.435 & 0.115 & 0.194 & 0.220 & 0.183 & 0.002 & 0.116 & 0.302 & 0.026 & 0.177 \\
        \midrule
        \multicolumn{12}{@{}l}{\emph{(c) Cross-axis excess degradation}} \\
        Human ID $\uparrow$ & $\Delta_{\mathrm{axis}}$ & +0.119 & -0.138 & +0.000 & +0.031 & +0.063 & +0.068 & +0.069 & -0.020 & -0.008 & \textbf{+0.020} \\
        Human App. $\uparrow$ & $\Delta_{\mathrm{axis}}$ & +0.193 & -0.021 & -0.010 & +0.079 & +0.058 & +0.040 & +0.185 & -0.073 & -0.013 & \textbf{+0.049} \\
        Object App. $\uparrow$ & $\Delta_{\mathrm{axis}}$ & +0.120 & +0.207 & +0.173 & +0.223 & +0.181 & +0.082 & +0.218 & +0.057 & -0.121 & \textbf{+0.127} \\
        Object Geometry $\uparrow$ & $\Delta_{\mathrm{axis}}$ & +0.156 & +0.158 & +0.181 & +0.231 & +0.117 & +0.096 & +0.182 & +0.035 & -0.080 & \textbf{+0.120} \\
        Env. Visible $\uparrow$ & $\Delta_{\mathrm{axis}}$ & +0.025 & +0.312 & +0.331 & +0.104 & +0.207 & +0.368 & +0.007 & +0.002 & +0.013 & \textbf{+0.152} \\
        Env. Depth $\uparrow$ & $\Delta_{\mathrm{axis}}$ & +0.020 & +0.220 & +0.191 & +0.021 & +0.113 & +0.465 & +0.008 & -0.098 & +0.026 & \textbf{+0.107} \\
        Env. Recon. $\uparrow$ & $\Delta_{\mathrm{axis}}$ & -0.027 & +0.320 & +0.017 & -0.149 & +0.097 & +0.395 & -0.099 & +0.053 & +0.004 & \textbf{+0.068} \\
        \bottomrule
    \end{tabular}%
    }
\end{table}
% END inlined sections/generated_cross_axis_revisit_table.tex

\subsection{Memory Evaluation under Departure and Return Controls}
\label{app:revisit-controls}

Inaccurate camera returns can reduce memory scores even when scene content
is preserved. We therefore test whether multi-revisit degradation and the
cross-axis penalty persist among videos with evidence of departure and
return. This analysis uses the existing evaluation results and leaves the
main benchmark scores unchanged.

\noindent\textbf{Controlling departure and return.}
For Human and Object, we retain videos in which the tracker provides
departure evidence and valid observations after return for every initially
evaluated target at every compared revisit. Targets that remain continuously
visible do not satisfy this criterion. For Environment, DA3 poses must
indicate a departure of at least $30^\circ$ from the initial orientation
and a return within $30^\circ$ of it. Tab.~\ref{tab:revisit-control-coverage}
reports the resulting coverage. These criteria use tracking and estimated
orientation as evidence of the executed trajectory; Reconstruction retains
the same zero-score rule for invalid revisits as the main evaluation.

\begin{table}[t]
\centering
\caption{Tracking-supported target revisits and pose-qualified Environment
returns. Entity columns are percentages averaged over targets, prescribed
revisits, and videos; Object uses the shared appearance/geometry tracking
evidence. Environment percentages use 490 slots per model.
$D$: departure $\geq30^\circ$; $R_{30}$: return error $\leq30^\circ$;
$Q_{30}$: departure $\geq30^\circ$ and return error $\leq30^\circ$.
The last columns count retained Environment model--case units for
multi-revisit ($N_{\mathrm{multi}}$) and cross-axis ($N_{\mathrm{cross}}$) comparisons.}
\label{tab:revisit-control-coverage}
\begingroup
\fontsize{8}{10}\selectfont
\setlength{\tabcolsep}{3pt}
\begin{tabular}{@{}lrrrrrrr@{}}
\toprule
& \multicolumn{3}{c}{Tracking support (\%)}
& \multicolumn{2}{c}{Environment pose (\%)}
& \multicolumn{2}{c}{$Q_{30}$ paired cases}\\
\cmidrule(lr){2-4}\cmidrule(lr){5-6}\cmidrule(lr){7-8}
Model & H.ID & H.App. & Object & $D$ & $R_{30}$
& $N_{\mathrm{multi}}$ & $N_{\mathrm{cross}}$\\
\midrule
HY-WorldPlay & 80.8 & 58.6 & 65.2 & 99.2 & 99.4 & 68 & 67 \\
Matrix-Game 3 & 56.7 & 45.1 & 31.7 & 99.4 & 91.2 & 69 & 25 \\
Matrix-Game 3.5 & 3.5 & 6.9 & 61.2 & 100.0 & 91.0 & 62 & 47 \\
SANA-WM & 26.6 & 34.7 & 43.9 & 99.6 & 80.8 & 54 & 25 \\
YUME & 30.0 & 19.5 & 22.0 & 92.9 & 94.5 & 52 & 41 \\
LingBot World & 34.7 & 36.3 & 45.0 & 98.6 & 80.0 & 41 & 1 \\
LingBot World v2 & 67.5 & 71.4 & 62.1 & 100.0 & 57.6 & 11 & 20 \\
Evoke & 84.1 & 77.6 & 78.3 & 99.6 & 95.3 & 65 & 52 \\
DreamX World & 27.6 & 19.5 & 15.6 & 76.9 & 58.4 & 9 & 3 \\
\bottomrule
\end{tabular}
\endgroup
\end{table}

Comparisons within each model use a fixed set of cases: all three returns
must qualify for LRLRLR, and both returns in both trajectories must qualify
for LRLR--LRUD. We compute each case's revisit decrease or cross-axis
difference before averaging over cases and then equally over models with
qualifying cases. The unfiltered comparison uses the same models.
Tab.~\ref{tab:revisit-controls} reports the retained sample sizes alongside
the effects, since some model-specific subsets contain only one or two cases.

\begin{table}[t]
\centering
\caption{Paired revisit diagnostics. $N$ counts retained model--case units
(a cross-axis unit contains two videos); $M$ counts contributing models.
All cases uses the same $M$ models as Qualified, without case filtering.
Positive reports the number of models with a positive conditional effect.
Tracking requires support for every target at every compared revisit;
Pose uses the $30^\circ$ criterion. Values are descriptive means, not
significance tests or conditional model rankings.}
\label{tab:revisit-controls}
\begingroup
\small
\setlength{\tabcolsep}{3pt}
\begin{tabular}{@{}lrrcccl@{}}
\toprule
Metric & $N$ & $M$ & All cases & Qualified & Positive & Criterion\\
\midrule
\multicolumn{7}{l}{\textit{(a) Multi-revisit decrease: $R_1-R_3$}} \\
Human Identity & 125 & 8 & 0.237 & 0.110 & 8/8 & Tracking \\
Human Appearance & 93 & 7 & 0.162 & 0.058 & 5/7 & Tracking \\
Object Appearance & 155 & 9 & 0.229 & 0.158 & 9/9 & Tracking \\
Object Geometry & 155 & 9 & 0.205 & 0.101 & 9/9 & Tracking \\
Environment Visible & 431 & 9 & 0.161 & 0.152 & 9/9 & Pose $30^\circ$ \\
Environment Depth & 431 & 9 & 0.138 & 0.135 & 8/9 & Pose $30^\circ$ \\
Environment Reconstruction & 431 & 9 & 0.196 & 0.139 & 9/9 & Pose $30^\circ$ \\
\midrule
\multicolumn{7}{l}{\textit{(b) Cross-axis excess degradation: $\Delta_{\mathrm{axis}}$}} \\
Human Identity & 135 & 8 & 0.023 & 0.034 & 6/8 & Tracking \\
Human Appearance & 109 & 8 & 0.056 & 0.050 & 6/8 & Tracking \\
Object Appearance & 141 & 8 & 0.117 & 0.023 & 4/8 & Tracking \\
Object Geometry & 141 & 8 & 0.115 & 0.051 & 6/8 & Tracking \\
Environment Visible & 281 & 9 & 0.152 & 0.128 & 8/9 & Pose $30^\circ$ \\
Environment Depth & 281 & 9 & 0.107 & 0.076 & 8/9 & Pose $30^\circ$ \\
Environment Reconstruction & 281 & 9 & 0.068 & 0.056 & 5/9 & Pose $30^\circ$ \\
\bottomrule
\end{tabular}
\endgroup
\end{table}

\noindent\textbf{Multi-revisit degradation persists after controlling return accuracy.}
Among the 431 qualifying Environment model--case pairs across all nine models,
Visible, Depth, and Reconstruction decrease from $R_1$ to $R_3$ by
0.152, 0.135, and 0.139, respectively. Visible and Reconstruction decline
for all nine models, while Depth declines for eight. These decreases are
smaller than the corresponding unfiltered values of 0.161, 0.138, and
0.196, but remain positive.
The tracking-supported subsets likewise show mean decreases in Human
Identity, Human Appearance, Object Appearance, and Object Geometry
(0.110, 0.058, 0.158, and 0.101). Thus, content recovery deteriorates across
revisits even among videos that satisfy the departure and return criteria.

\noindent\textbf{The average cross-axis penalty persists, with greater variation across models.}
On 281 qualifying case pairs across all nine models, the mean additional
degradation under LRUD is 0.128 for Environment Visible, 0.076 for Depth,
and 0.056 for Reconstruction. The effect is positive for eight of nine
models on Visible and Depth, but five of nine on Reconstruction.
All four entity metrics also retain a positive mean cross-axis penalty;
Object Appearance, however, has four positive and four negative model-level
effects. Switching axes therefore remains harder on average in these
subsets, although the effect is less consistent across models than
multi-revisit degradation.

\begin{table}[t]
\centering
\caption{Environment return-tolerance sensitivity, always requiring
departure $\geq30^\circ$ and fixed cases across compared revisits.
$N/M$ denotes retained model--case units / contributing models.
Vis., Dep., and Rec.\ denote Visible, Depth, and Reconstruction.
The retained case and model sets change with the tolerance.}
\label{tab:revisit-pose-sensitivity}
\begingroup
\small
\setlength{\tabcolsep}{3pt}
\begin{tabular}{@{}crrrr rrrr@{}}
\toprule
& \multicolumn{4}{c}{Multi-revisit decrease}
& \multicolumn{4}{c}{Cross-axis excess degradation}\\
\cmidrule(lr){2-5}\cmidrule(lr){6-9}
Tolerance & $N/M$ & Vis. & Dep. & Rec. & $N/M$ & Vis. & Dep. & Rec.\\
\midrule
$5^\circ$ & 195/8 & 0.092 & 0.057 & 0.116 & 30/4 & 0.094 & 0.158 & 0.050 \\
$15^\circ$ & 345/8 & 0.116 & 0.085 & 0.141 & 134/7 & 0.090 & 0.010 & 0.029 \\
$30^\circ$ & 431/9 & 0.152 & 0.135 & 0.139 & 281/9 & 0.128 & 0.076 & 0.056 \\
\bottomrule
\end{tabular}
\endgroup
\end{table}

As stricter checks of the $30^\circ$ analysis, we repeat the comparisons
with return tolerances of $15^\circ$ and $5^\circ$. Both mean effects
remain positive for all three Environment metrics
(Tab.~\ref{tab:revisit-pose-sensitivity}).
Tighter tolerances reduce coverage: the $5^\circ$ cross-axis comparison
contains only 30 case pairs from four models, so effect sizes across
tolerances also reflect changes in the evaluated samples.
Together, these results show that the degradation is not confined to
videos with poorly executed returns. Residual differences in camera
position, occlusion, and estimation accuracy can still affect the scores;
the analysis supports the observed trends rather than attributing the
remaining decrease exclusively to forgetting.
\FloatBarrier

\section{Analysis of Context-Access Policies}
\label{app:context-policy-analysis}
In this section, we examine the capability differences induced by temporal
KV-access policies in Section~\ref{sec:lingbot-v2-context-management},
detailing the controlled setup and the accompanying capability analysis.

\noindent\textbf{Controlled Experimental Setup.}
We compare FullKV, Sliding Windows, and FoV on LingBot World v2 without
retraining. We fix the checkpoint, conditioning first frame, text prompt,
camera trajectory, random seed, number of sampling steps, generation length,
and evaluators. The temporal KV-access policy during autoregressive generation is the only variable. FullKV allows the current chunk to access the entire historical KV context.
Sliding Windows uses a fixed budget of 20 latents: 4 sink, 12 recent, and
4 current latents. FoV uses the same 20-latent budget, comprising 4 sink,
4 recent, 8 retrieved, and 4 current latents. The retrieved latents come from
two historical chunks selected by current-to-history field-of-view overlap.
The fixed-budget comparison therefore tests how allocating context to
view-relevant history affects memory beyond access to recent observations.

\noindent\textbf{Evaluation coverage and timing.} All three policies use identical evaluation sets: 180 Human, 210 Object,
210 Environment, and 60 Dynamic Memory videos, totaling 660 matched videos
per policy. Latency is the mean end-to-end time to generate a 493-frame video
over this matched set. Speedup is the ratio of FullKV's mean latency to the
corresponding policy's mean latency.

\begin{figure}[h]
    \centering
    \vspace{-0.15in}
    \includegraphics[width=\textwidth]{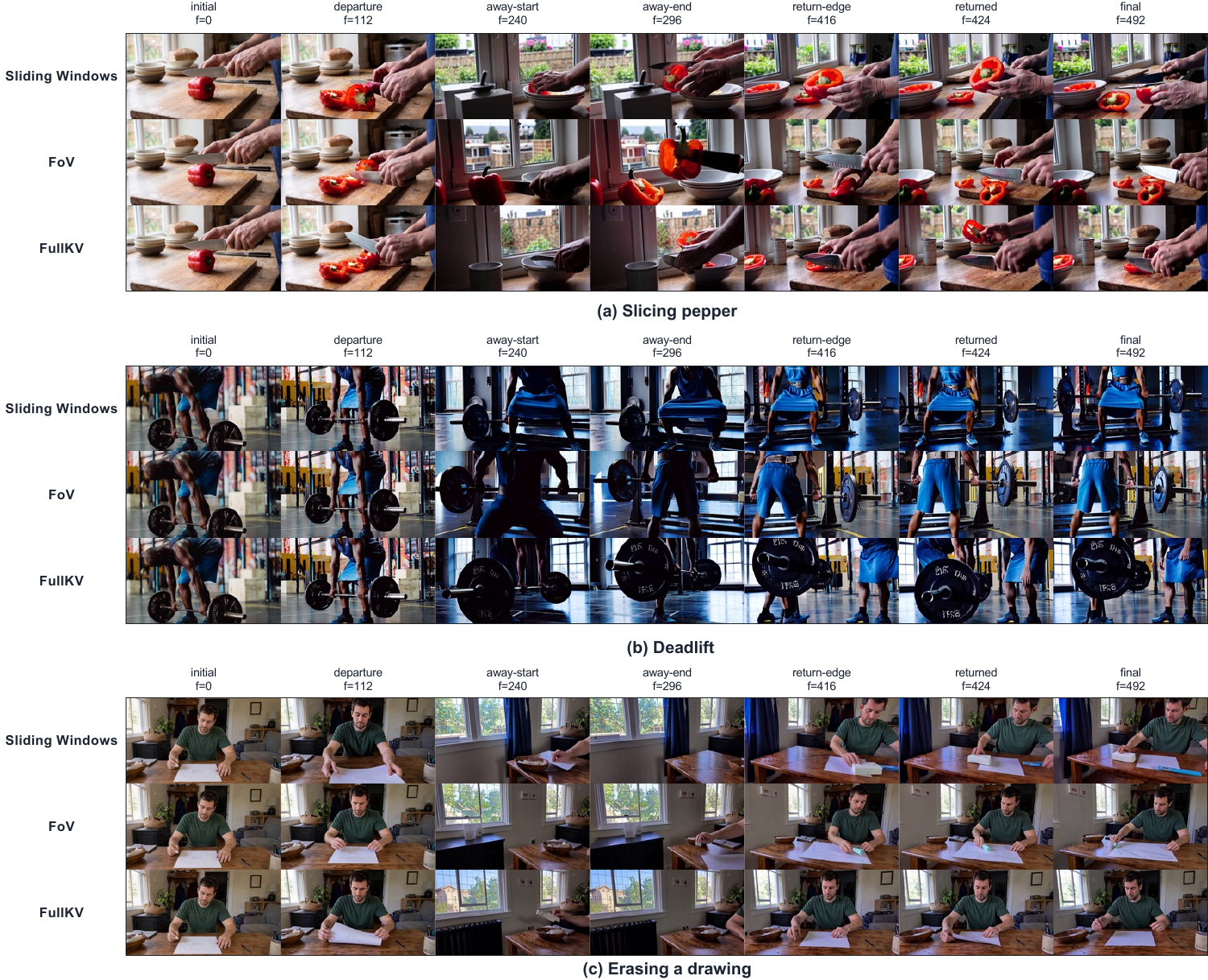}
    \vspace{-0.25in}
    \caption{\textbf{World memory under different context-access policies
    in LingBot World v2.}
    The examples depict (a) slicing a pepper, (b) performing a deadlift,
    and (c) erasing a drawing.
    Within each example, rows show Sliding Windows, FoV, and FullKV
    from top to bottom.
    Columns present selected frames along the prescribed LR-Hold
    trajectory, spanning the initial observation, departure, and revisit.
    Camera phases and frame indices are indicated above.}
    \label{fig:lingbot-v2-dynamic-context}
\end{figure}

\noindent\textbf{View-relevant context for static retention.}
Across 490 Environment revisit slots, FoV increases the proportion
of returns that reach the target viewpoint from 57.6\% to 65.3\%
and reduces the mean return rotation error from $31.0^\circ$
to $26.7^\circ$ relative to FullKV.
These gains accompany improvements in Human Identity, Human Appearance,
Environment Depth, and Environment Reconstruction, suggesting that
view-relevant historical context supports both scene recovery and
content retention.
The deadlift example in Fig.~\ref{fig:lingbot-v2-dynamic-context}
illustrates the difference in viewpoint recovery: FoV returns to
a more complete view of the lifter and barbell, whereas FullKV
retains a close-up dominated by the weight plates.

\noindent\textbf{Context allocation for dynamic progression.}
Maintaining dynamic memory requires the model to preserve the latest
event state while allowing it to evolve.
Sliding Windows matches FullKV on Return Progress (0.667), improves
Final State from 0.592 to 0.600, but scores lower on Persistence/Reset
(0.717 vs.\ 0.733) and Causal/Physics (0.775 vs.\ 0.792).
Prioritizing recent observations alone therefore does not consistently
improve dynamic memory.
FoV combines recent context with view-relevant historical evidence
and achieves further gains over both baselines across all four
sub-metrics.
The erasing example in Fig.~\ref{fig:lingbot-v2-dynamic-context}
also reveals differences in the consistency of task-relevant objects:
Sliding Windows produces a visibly enlarged erasing object,
while FoV maintains a compact eraser in contact with the paper
during the returned sequence.
Together, the quantitative comparison and these examples favor combining
recent states with view-relevant historical context over a recent-only window.

\section{Limitation and Future Work}
\label{app:limitation}
The current benchmark focuses on controlled translation and rotation
trajectories over rollouts of several hundred frames. Longer horizons,
free-form or compound traversal patterns, and embodied settings in which camera
and object actions are jointly selected remain outside its present scope.
Dynamic Memory additionally relies on frozen case-specific checklists evaluated
by a VLM. Although the resulting scores show strong overall agreement with
human judgments, the Causal/Physics dimension remains the most challenging
sub-dimension. Extending the benchmark to richer event types, multi-agent
interactions, and longer unobserved phases, together with stronger and more
diverse evaluators, would further test the limits of persistent world
generation. We hope \benchname{} serves as a practical testbed for developing
video world models that remember, evolve, and remain consistent over extended
interaction.

\end{document}

%% file: math_commands.tex
\usepackage{amsmath,amsfonts,bm}

\def\eqref#1{equation~\ref{#1}}
\def\1{\bm{1}}

\DeclareMathAlphabet{\mathsfit}{\encodingdefault}{\sfdefault}{m}{sl}
\SetMathAlphabet{\mathsfit}{bold}{\encodingdefault}{\sfdefault}{bx}{n}